%% file: icml2026_paper.tex
\documentclass{article}

\usepackage{microtype}
\usepackage{graphicx}
\usepackage{subcaption}
\usepackage{booktabs} 
\usepackage{algorithm}
\usepackage{algorithmic}
\usepackage{tabularx}
\usepackage{adjustbox}
\usepackage{multirow}
\usepackage{array}
\usepackage[table]{xcolor}
\usepackage{wrapfig}
\usepackage{float}
\usepackage{enumitem}
\usepackage{pifont}
\newcommand{\xmark}{\ding{55}}

\usepackage{bbm} 
\usepackage[most]{tcolorbox}
\usepackage{xcolor}
\usepackage{hyperref}
\usepackage[T1]{fontenc}
\renewcommand{\xmark}{\ding{55}} 
\usepackage{enumitem}
\newcommand{\methodname}{{\textsc{AnomSeer}}}
\usepackage{bbm}
\usepackage{amssymb}
\usepackage[dvipsnames]{xcolor}
\usepackage[T1]{fontenc}
\definecolor{midnightgreen}{rgb}{0.0, 0.29, 0.33}
\definecolor{deepgreen}{HTML}{055c29}
\definecolor{deeppurple}{HTML}{7030a0}
\definecolor{deepblue}{HTML}{171d91}
\definecolor{brown}{HTML}{843c0c}
\definecolor{shadered}{HTML}{ffe5e5}
\definecolor{shadegreen}{HTML}{e5f7ed}
\definecolor{msftBlack}{RGB}{0,0,0}
\definecolor{lightred}{RGB}{255,163,163}
\definecolor{deepred}{RGB}{153,0,0}

\definecolor{barblue}{RGB}{90,120,180}
\definecolor{barorange}{RGB}{225,124,5}

\definecolor{softblue}{RGB}{30, 90, 160}
\hypersetup{
  colorlinks=true,
  linkcolor=deepred,
  citecolor=brown,
  urlcolor=magenta
}

\definecolor{blueviolet}{RGB}{138,43,226}
\newtcolorbox{insightblock}{
  colback=blueviolet!5,   
  colframe=blueviolet!50!black!50!,    
  boxrule=0.5mm,       
  arc=1mm,             
  left=2pt,            
  right=2pt,           
  top=1pt,             
  bottom=1pt,          
  after skip=5pt,
  before skip=5pt,
}
\input{math_commands.tex}

\newtcolorbox{AIbox}[1][]{
    title={#1},
    colback=gray!10,
    colframe=black
}

\newtcolorbox{AIboxDark}[1][]{
    title={#1},
    colback=yellow!20!brown!10,
    colframe=orange!60!black,
    coltitle=white,
    colbacktitle=orange!70!black
}
\usepackage{hyperref}

\usepackage[accepted]{icml2026}

\usepackage{amsmath}
\usepackage{amssymb}
\usepackage{mathtools}
\usepackage{amsthm}

\usepackage[capitalize,noabbrev]{cleveref}

\theoremstyle{plain}

\theoremstyle{definition}

\theoremstyle{remark}

\usepackage[textsize=tiny]{todonotes}

\icmltitlerunning{AnomSeer: Reinforcing Multimodal LLMs to Reason for Time-Series Anomaly
Detection}

\begin{document}

\twocolumn[
  \icmltitle{AnomSeer: Reinforcing Multimodal LLMs to Reason for Time-Series Anomaly Detection}



  \icmlsetsymbol{equal}{*}

\begin{icmlauthorlist}
  \icmlauthor{Junru Zhang}{zju}
  \icmlauthor{Lang Feng}{ntu}
  \icmlauthor{Haoran Shi}{ntu}
  \icmlauthor{Xu Guo}{ntu}
  \icmlauthor{Han Yu}{ntu}
  \icmlauthor{Yabo Dong}{zju}
  \icmlauthor{Duanqing Xu}{zju}
\end{icmlauthorlist}

\icmlaffiliation{zju}{Zhejiang University}
\icmlaffiliation{ntu}{Nanyang Technological University}

\icmlcorrespondingauthor{Xu Guo}{xu.guo@ntu.edu.sg}
\icmlcorrespondingauthor{Yabo Dong}{dongyb@zju.edu.cn}

\icmlkeywords{Machine Learning, ICML}

\vskip 0.3in
]


\printAffiliationsAndNotice{}  
\newcommand{\fix}{\marginpar{FIX}}
\newcommand{\new}{\marginpar{NEW}}


\begin{abstract}
Time-series anomaly detection (TSAD) with multimodal large language models (MLLMs) is an emerging area, yet a persistent challenge remains: MLLMs rely on coarse time-series heuristics but struggle with multi-dimensional, detailed reasoning, which is vital for understanding complex time-series data. We present \methodname{} to address this by reinforcing the model to ground its reasoning in precise, structural details of time series, unifying anomaly classification, localization, and explanation.
    At its core, an expert chain-of-thought trace  is generated to provide verifiable, fine-grained reasoning from classical analyses (e.g., statistical measures, frequency transforms). Building on this, we propose a novel \underline{time}-series g\underline{r}ounded \underline{p}olicy \underline{o}ptimization (TimerPO) that incorporates two additional components beyond standard reinforcement learning: a time-series grounded advantage based on optimal transport and an orthogonal projection to ensure this auxiliary granular signal does not interfere with the primary detection objective. Across diverse anomaly scenarios, \methodname{}, with Qwen2.5-VL-3B/7B-Instruct, outperforms larger commercial baselines  in classification and localization accuracy, particularly on point- and frequency-driven exceptions. Moreover, it produces plausible reasoning traces that support its conclusions.

\end{abstract}
\begin{figure*}[h]
    \centering
    \includegraphics[width=0.92\textwidth]{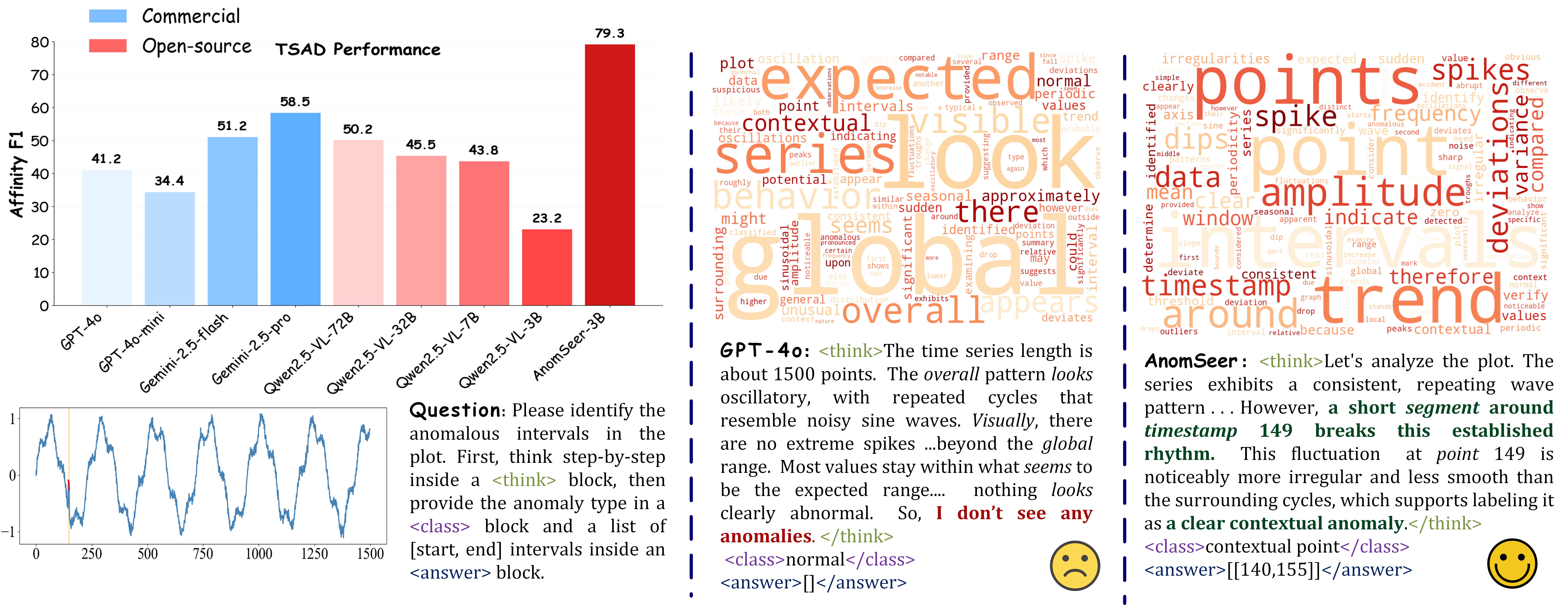}
 
    \caption{Comparison of model performance and time-series reasoning quality. \textbf{Left}: Affinity F1 (\%) of different models on TSAD benchmarks. \textbf{Middle}: GPT-4o results, including word frequency distributions in reasoning (top) and its coarse-grained answer (bottom). \textbf{Right}: \methodname{} results, including word frequency distributions in reasoning (top) and its fine-grained answer (bottom).}
    \label{fig:example}
    
\end{figure*}

\section{Introduction}

Recent advances in large language models (LLMs) have opened new opportunities for time-series anomaly detection (TSAD)~\citep{xu2021anomaly}. Building on this progress, we focus on a practical yet underexplored setting, \emph{time-series reasoning for anomalies}~\citep{yang2025time,kong2026achievingtimeseriesreasoning}, where the goal goes beyond flagging abnormal segments: models must also produce coherent, linguistically grounded explanations. Emerging studies~\citep{zhou2024can,xu2025can,he2025harnessing} have revealed that LLMs exhibit stronger zero-shot robustness when reasoning over visual renderings of time series (e.g., line plots) rather than raw numeric sequences. This advantage arises from human-like pattern perception and greater token efficiency enabled by compact, semantically rich images~\citep{he2025harnessing,liu2024picture}. These insights naturally motivate multimodal LLMs (MLLMs) as the backbone for advancing TSAD in a \emph{reasoning-centric} manner, i.e., detecting, attributing, and justifying anomalies with structured natural language grounded in visual cues. 

Despite these strengths, MLLMs fundamentally lack built-in time-series priors, and their reasoning often resorts to coarse time-series heuristics and struggles with detailed time-series analysis (Figure~\ref{fig:example} (Middle)), thereby leading to suboptimal performance. While reinforcement learning (RL)~\citep{sutton2018reinforcement} has proven more effective than supervised fine-tuning (SFT)~\citep{zhang2025timemaster,liu2025time,luo2025time,tan2025inferring} at incentivizing the emergent reasoning of LLMs in other domains~\citep{guo2025deepseek,wei2025swe,feng2025group}, its reliance on globally verifiable rule-based goals may be ill-suited for the model to capture subtle, fine-grained time-series patterns. Consequently, even well-trained MLLMs may only excel at salient, out-of-range anomalies yet struggle to articulate nuanced shifts (e.g., small trend drifts) with faithful textual evidence. This discrepancy raises a central question for MLLMs in TSAD:

 \vspace{5pt}
\begin{tcolorbox}[
notitle, 
rounded corners, 
colframe=darkgray, 
colback=white, 
  boxrule=0.5mm,       
  arc=1mm,             
  left=2pt,            
  right=2pt,           
  top=4pt,             
  bottom=4pt,          
  after skip=5pt,
  before skip=5pt,
]
\emph{
  {
    \centering 
  {
Can we incentivize MLLMs to ground their time-series reasoning in fine-grained, multi-dimensional evidence, ensuring faithful and verifiable anomaly interpretations?
  }
  \\
  }
}
 \vspace{5pt}
\end{tcolorbox}

To address this challenge, we propose \methodname{}\footnote{\url{https://github.com/jrzhang33/AnomSeer}.}, a novel time-series MLLM post-training approach that not only detects anomalies but also produces structured, evidence-based explanations to support its decisions. Our core idea is to fuse the analytical rigor of classical numerical TSAD with the holistic visual intuition of MLLMs through two components:
\emph{(i) expert chain-of-thought (ExpCoT)} trace, which encodes structured reasoning inspired by classical TSAD workflows, and  
\emph{(ii) \underline{time}-series g\underline{r}ounded \underline{p}olicy \underline{o}ptimization (TimerPO)}, a novel temporal-aware RL algorithm that softly aligns the model’s reasoning with ExpCoT trajectories.
Instead of merely correcting outputs, \methodname{} utilizes the analytical rigor of traditional TSAD methods, such as residual inspection~\citep{hyndman2018forecasting} and wavelet-based drift detection~\citep{thill2017time}, and embeds it into the MLLM's learning process. TimerPO operationalizes this integration by measuring the semantic deviation from an ExpCoT using optimal transport~\citep{caffarelli2010free,bonneel2011displacement} and transforms this distance into a refinement advantage signal. This signal is then orthogonally projected, ensuring it acts as non-interfering auxiliary guidance of the main RL objective. Consequently, TimerPO enhances the model's fine-grained temporal-aware reasoning capabilities (Figure~\ref{fig:example} (Right)) without perturbing its global visual understanding or the primary optimization objective. We summarize our key contributions as follows:
\noindent
\begin{itemize}[leftmargin=*]
\item We explore a pivotal challenge hindering the effectiveness of MLLMs for TSAD: the tendency of MLLMs to rely on coarse visual ``eyeballing'' rather than engaging in fine-grained numerical reasoning. We introduce \methodname{}, a novel approach that bridges this gap by transferring classical, detailed TSAD priors into the time-series reasoning process of MLLMs during training.
\item  We propose TimerPO, a new RL algorithm designed for time-series reasoning in TSAD. TimerPO guides fine-grained, numerical time-series knowledge into the model's reasoning.  It leverages optimal transport to create auxiliary advantage signals and applies them as non-interfering corrective guidance for RL training via orthogonal projection.
\item Extensive experiments across diverse TSAD tasks demonstrate that \methodname{} consistently outperforms strong MLLM baselines (e.g., GPT-4o) in detection accuracy and localization precision, unifying detection, categorization, and reasoning. Critically, it produces fine-grained, plausible reasoning traces grounded in detailed time-series evidence, achieving faithful and verifiable interpretations in time-series anomaly detection.
\end{itemize}

\section{Related Work}
\textbf{Time series anomaly detection} is a critical task in domains like healthcare, aiming to detect deviations from normal temporal patterns~\citep{wu2024catch,shentu2024towards}. Traditional methods rely on statistical techniques and machine learning methods (e.g., Z-score~\citep{bhatnagar2021merlion}, Isolation Forest~\citep{liu2008isolation} and One-Class SVM~\citep{scholkopf1999support}), while recent advances use deep models such as Autoencoders~\citep{zong2018deep,park2018multimodal} for reconstruction- or prediction-based detection. Despite their effectiveness, these models struggle in industrial settings due to the scarcity of anomaly data, limiting generalization. To address this, recent efforts explore pre-trained~\citep{zhou2023one, zhang2025timesbert} and time-series foundation models~\citep{goswami2024moment, gao2024units}  for zero- and few-shot detection. However, these approaches are primarily optimized for accuracy, \emph{lacking} the ability to analyze anomaly types, reason about temporal patterns, and explain why a given sample is anomalous. 

\textbf{Time-series reasoning with LLMs} is an emerging research frontier~\citep{kong2026achievingtimeseriesreasoning}.
To enable LLMs to perform time-series analysis, researchers have primarily explored two input strategies:
prompting with numerical data~\citep{alnegheimish2024large} or visual representations~\citep{zhuang2024see,he2025harnessing, xu2025can,zhou2024can}.
While the visual approach, feeding plots into MLLMs such as GPT-4o, is often more token-efficient, its effectiveness is limited by the fact that these models are not explicitly trained on time-series visualizations.
To instill temporal understanding, recent works have primarily relied on integrating classical modules~\citep{chen2025synergizing,liu2025large}, employing auxiliary techniques~\citep{he2025harnessing,zhuang2024see}, or large-scale SFT~\citep{yang2025time}.
An alternative and promising path involves RL to promote structured problem-solving, as seen in DeepSeek-R1~\citep{guo2025deepseek}. Building on this, recent work such as TimeMaster~\citep{zhang2025timemaster} trains MLLMs for classification tasks by combining SFT with RL to enable interpretable temporal reasoning over visualized series.
Nevertheless, RL for enhancing anomaly detection in MLLMs remains underexplored.
In this paper, we show that vanilla RL struggles to detect subtle anomalies and propose a new method to mitigate this limitation. 


\section{Preliminary}
\textbf{Time-series anomaly detection.} Time-series anomaly detection (TSAD) aims to identify abnormal patterns within temporal data. 
Following standard practice~\citep{zhou2024can}, we use $\mathbf{X} = \{ \mathbf{x}_t \}_{t=1}^{T}$ to denote a univariate time series of length $T$, where each observation $\mathbf{x}_t \in \mathbb{R}$ is sampled at regular intervals and may correspond to either normal or anomalous behavior.
Anomalies are defined as continuous intervals of data points that deviate significantly from the expected pattern. They can be categorized into point-wise anomalies (contextual point and global point) and range-wise anomalies (trend, shapelet, and seasonal), resulting in five types in total.
Formally, let $\mathcal{A} = \{ (t_s^{(i)}, t_e^{(i)}) \}_{i=1}^{k}$ denote the set of anomalous intervals, where $1 \leq t_s^{(i)} \leq t_e^{(i)} \leq T$. 
Each tuple $(t_s^{(i)}, t_e^{(i)})$ specifies the start and end indices of the $i$-th anomalous segment; in particular,  $t_s^{(i)} = t_e^{(i)}$ denotes a single-point anomaly. 
The primary goal of TSAD is to infer the set $\mathcal{A}$ with high accuracy. 

\textbf{Multimodal time-series formulation.} To enable MLLMs to perform time-series anomaly detection, the input of the MLLM consists of the time-series input $\mathbf{X}$ and context prompt $\mathbf{c}$ that encodes domain knowledge, natural-language instructions, or task-specific queries to guide the model’s reasoning process.
To enable multimodal processing, we follow the \emph{visualization input strategy}~\citep{liu2024picture,xu2025can,zhang2025timemaster}, rendering the raw time series into a line-plot image $\mathbf{X} \xrightarrow{} \mathbf{I}$ and then feeding it to the MLLM’s vision encoder. 
This approach allows the model to leverage its pre-trained visual reasoning abilities on a representation that is both compact and semantically rich~\citep{xu2025can,xie2024chatts}.

\textbf{Multimodal LLM inference.}
We define a time-series MLLM $\pi_\theta$ (parameterized by $\theta$) that specifies a conditional distribution over an output sequence $\mathbf{y}=\{y_1,y_2,\dots,y_N\}$, where each token $y_n$ may correspond to an anomaly label, an interval boundary, or a natural-language reasoning.
Given the rendered time-series data $\mathbf{I}$ and textual context $\mathbf{c}$, the model generates outputs autoregressively: $\pi_{\theta}(\mathbf{y}\mid \mathbf{I}, \mathbf{c})=\prod_{n=1}^{N} \pi_{\theta}\!\left(y_n \mid y_{<n},\, \mathbf{I},\, \mathbf{c}\right)$.
This formulation unifies reasoning, explanation and detection in a single generative process, allowing the model to produce structured outputs that are both context-aware and interpretable.

\begin{figure*}[t]
\centering
\includegraphics[width=0.95\textwidth]{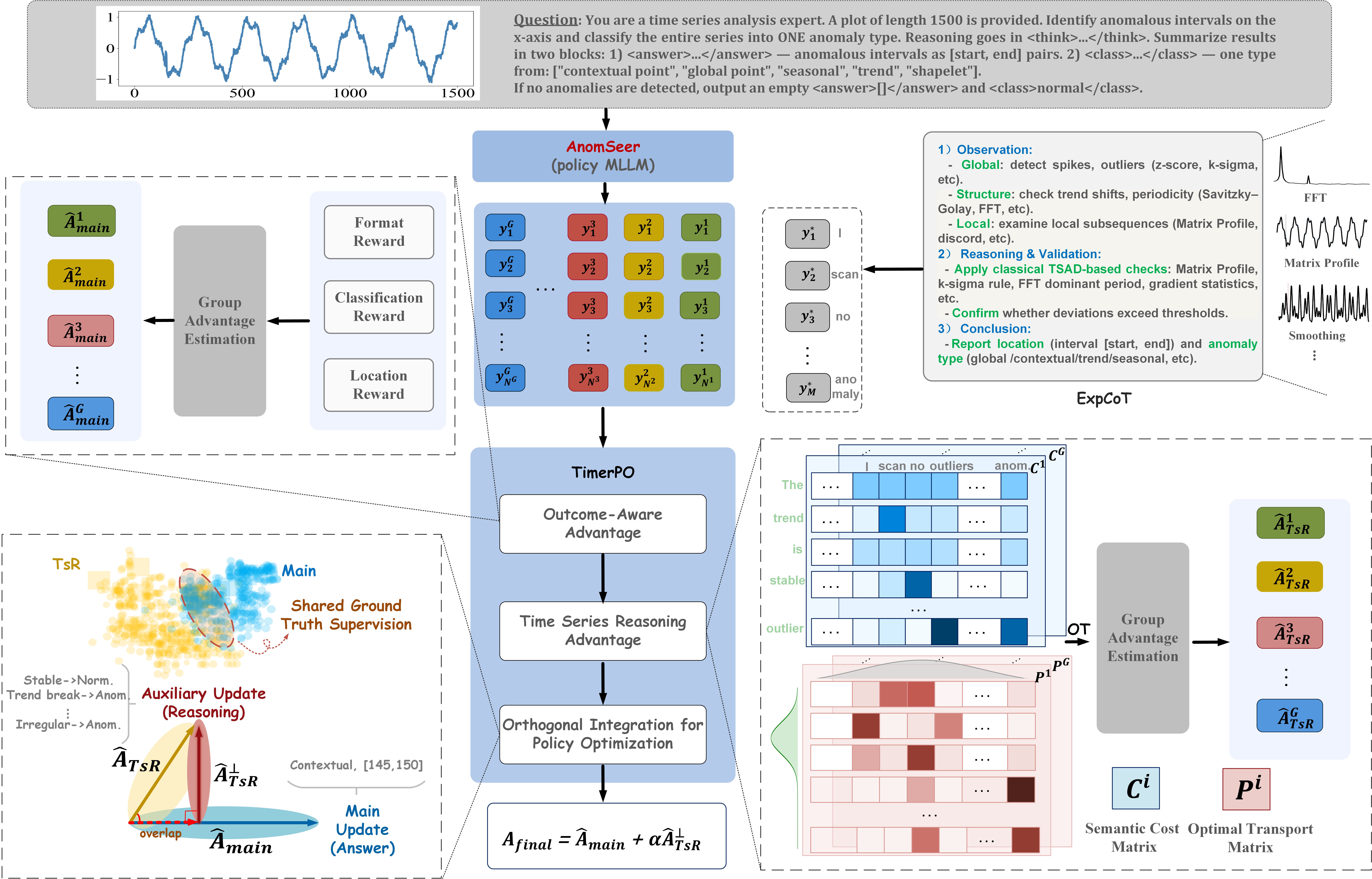}
\caption{The overall framework of \methodname{}. \methodname{} first generates ExpCoT reasoning traces $\mathbf{y}^*$ from the time-series data based on classical TSAD techniques (e.g., FFT). TimerPO then computes the outcome-aware advantage and leverages optimal transport to compute the time-series reasoning advantage, which is orthogonally integrated into policy optimization to ensure stable training and improved reasoning quality.}
\label{fig:method}

\end{figure*}
\section{Methodology}
Time-series MLLMs often rely on coarse visual heuristics and fail to produce numerically grounded, fine-grained reasoning for TSAD. This weakness limits their ability to detect subtle anomalies such as frequency shifts or small trend drifts in complex time-series data. To address this, we introduce \methodname{}, a novel MLLM post-training approach for TSAD that couples classical time-series statistical rigor with the expressive reasoning ability of MLLMs. \methodname{} is trained with two key components: 
(1) \emph{expert chain-of-thought (ExpCoT)}, which generates structured, expert-like reasoning traces from ground-truth time series using statistical diagnostics (e.g., histogram-based outlier scores, FFT, matrix profile); and 
(2) \emph{\underline{time}-series g\underline{r}ounded \underline{p}olicy \underline{o}ptimization (TimerPO)}, a new RL algorithm that leverages ExpCoT to establish the corrective, orthogonal advantages to refine reasoning without overriding the detection objective. 
Figure~\ref{fig:method} presents an overview of \methodname{}. In the remainder of this section,
we will detail the design of ExpCoT (Section~\ref{sec:expcot}) and the TimerPO optimization algorithm (Section~\ref{sec:timerpo}), and discuss how they jointly enable accurate, interpretable, and numerically faithful anomaly detection.

\subsection{Expert Chain-of-Thought Generation}\label{sec:expcot}
To ground the reasoning of time-series MLLM with classical time-series detailed analysis for TSAD, we introduce the \emph{expert chain-of-thought (ExpCoT)} trace, a structured reasoning that mirrors the stepwise detection of a human analyst. Unlike an LLM-generated CoT, which may rely on heuristic pattern matching, ExpCoT is grounded in systematically derived, quantitatively verifiable evidence.
ExpCoT is generated per instance, starting from ground-truth annotations. We apply classical statistical and signal-processing techniques to extract descriptive statistics, candidate anomaly categories, and precise temporal localization. This trace delivers rich, multi-dimensional guidance that goes beyond a simple correct/incorrect signal, encouraging fine-grained and interpretable reasoning.

Crucially, ExpCoT adheres to a disciplined \emph{three-stage} reasoning path (\emph{Observation $\rightarrow$ Reasoning \& Validation $\rightarrow$ Conclusion}), closely mirroring the stepwise process of human analytical reasoning.

\begin{insightblock}
\textbf{Observation}: systematically progress from global patterns to local irregularities across multiple views, surfacing candidate anomalies prior to formal testing.
\end{insightblock}
The \emph{``Observation''} stage performs a hierarchical scan of the time series $\mathbf{X}$ to extract preliminary statistical features.
 (\textbf{1}) \emph{Global Scan:} We first assess extreme values by examining the global data distribution via a histogram-based outlier score~\citep{goldstein2012histogram}.
(\textbf{2}) \emph{Structural Scan:} If no global outliers are present, we analyze fundamental properties such as trend stability using smoothed gradients~\citep{thill2017time} and periodicity via FFT-based frequency analysis~\citep{ren2019time}.
(\textbf{3}) \emph{Local Scan:} If the series appears structurally stable, we perform a localized search for dissimilar subsequences (discords) using the Matrix Profile~\citep{yeh2016matrix}.
This fine-grained scan provides the key statistical features 
that guide the subsequent detection process.

\begin{insightblock}
\textbf{Reasoning \& Validation}: 
use classical TSAD techniques to formalize ground truth as fine-grained, testable claims, supported by targeted quantitative time-series analysis.
\end{insightblock}
The \emph{``Reasoning \& Validation''} stage establishes a causal link between preliminary observations and formal statistical evidence of anomalies. First, it leverages the ground-truth anomaly type to align statistical markers with visual patterns (e.g., {``A sharp spike around $t \approx 150$ deviates significantly from the rest of the data, suggesting a contextual anomaly''}). This classification then guides the selection of a targeted statistical method for validation; for example, a suspected trend shift is validated using gradient analysis~\citep{thill2017time}, while the aforementioned contextual anomaly is confirmed by its Matrix Profile score~\citep{yeh2016matrix}. The numerical outcome is translated into a natural language explanation (e.g., {``The discord's z-score of 4.2 at timestamp 145 exceeds the 3-sigma threshold, confirming a significant pattern deviation''}). 

\begin{insightblock}
\textbf{Conclusion}: integrate multi-dimensional insights and fine-grained evidence into a precise, defensible anomaly judgment.
\end{insightblock}
The final \emph{``Conclusion''} stage synthesizes the findings into a conclusive summary. It integrates the multi-dimensional understanding from the \emph{``Observation''} stage with the detailed, quantitative evidence from the \emph{``Reasoning \& Validation''} stage to deliver a definitive judgment, e.g., {``Therefore, the detected anomaly is a contextual point, located in the interval [145, 150]''}.

In summary, as shown in Figure~\ref{fig:method}, ExpCoT provides a structured reasoning trace that embeds analytical rigor and numerically grounded logic. This makes it particularly effective for identifying subtle anomalies and offers fine-grained, informed guidance for subsequent MLLM training. See examples of ExpCoT in Appendix~\ref{app:details_expcot}.

\subsection{Time-Series Grounded Policy Optimization}\label{sec:timerpo}
To leverage ExpCoT and enable the reasoning of MLLM grounded in fine-grained time-series analysis, we introduce TimerPO, a novel RL method building upon Group Relative Policy Optimization (GRPO)~\citep{shao2024deepseekmath}. We begin with the vanilla GRPO formulation. Given the rendered time-series instance $\mathbf{I}$ and textual context $\mathbf{c}$, the model produces a \emph{group} of candidate responses $\mathcal{G} = \{\mathbf{y}^{1},\mathbf{y}^{2},...,\mathbf{y}^{G}\}$ where $G$ denotes the group size. This group-based generation enables pairwise relative reward comparisons, which are subsequently used to compute group-aware advantages.

\paragraph{Outcome-Aware Advantage.}
For each generated response $\mathbf{y}^{i}\in\mathcal{G}$, the task reward is a weighted sum of
(i) a format reward $r^{\mathrm{fmt},\,i}\in\{0,1\}$ that checks if the predefined output format of time-series MLLM is valid,
(ii) a classification reward $r^{\mathrm{cls},\,i}$ for anomaly type accuracy
and (iii) a detection location reward $r^{\mathrm{loc},i}$, which integrates common anomaly-detection metrics~\citep{zhou2024can}:
\begin{equation}
r^{i} = \lambda^{\mathrm{fmt}} r^{\mathrm{fmt},\,i} \;+\; \lambda^{\mathrm{cls}} r^{\mathrm{cls},\,i} \;+\; \lambda^{\mathrm{loc}} r^{\mathrm{loc},\,i},
\end{equation}
where $\lambda^{\mathrm{fmt}}, \lambda^{\mathrm{cls}}, \lambda^{\mathrm{loc}}$ are tunable weights.
To stabilize optimization, rewards are normalized within each group, yielding the main advantage:
\begin{equation}
\label{eq:main_advantage_hat}
\widehat{A}_{\mathrm{main}}^{i}
=
\frac{r^{i} - \mu_r}{\sigma_r + \varepsilon},
\quad
\mu_r=\frac{1}{G}\sum\nolimits_{i=1}^{G} r^{i},
\end{equation}
where
$\sigma_r^2=\frac{1}{G}\sum\nolimits_{i=1}^{G}\big(r^{i}-\mu_r\big)^2$.
The vectorized form
$\widehat{A}_{\mathrm{main}}=(\widehat{A}_{\mathrm{main}}^{1},\dots,\widehat{A}_{\mathrm{main}}^{G})^\top\in\mathbb{R}^G$
serves as the normalized baseline signal for subsequent policy updates. However, such outcome-aware advantages risk encouraging coarse, heuristic reasoning for time series data (e.g., detecting only obvious outliers while ignoring subtle but meaningful temporal patterns).

\paragraph{Time-Series Reasoning Advantage.}
To explicitly encourage fine-grained reasoning, TimerPO leverages the Optimal Transport (OT)~\citep{villani2008optimal,li2024gilot} to quantify the semantic alignment between a model's reasoning trace $\mathbf{y}^{i}=\{y^{i}_1,\dots,y^{i}_{N^i}\}$ and the corresponding ExpCoT's reasoning trace $\mathbf{y}^{\star}=\{y^{\star}_1,\dots,y^{\star}_{M}\}$ where $N^i$ and $M$ are their lengths. Given $\mathbf{y}^{i}$ and $\mathbf{y}^{\star}$, we extract the final-layer embeddings from the MLLM $\pi_\theta$, obtaining embedding vectors $\mathbf{e}^{i}$ for $\mathbf{y}^{i}$ and $\mathbf{e}^{\star}$ for $\mathbf{y}^{\star}$. We then construct a semantic cost matrix $\mathbf{C}^{i}\in\mathbb{R}^{N^i\times M}$ whose $(n,m)$-th entry measures the cosine distance between token embeddings:
\begin{equation}
\small
C^{i}_{nm}=1-\frac{\mathbf{e}^{i}_n\!\cdot\!\mathbf{e}^{\star}_m}
{\lVert\mathbf{e}^{i}_n\rVert\,\lVert\mathbf{e}^{\star}_m\rVert},
\; n=1,\dots,N^i,\; m=1,\dots,M.
\end{equation}
Let $\mathbf{u}^{i}\in\Delta^{N^i-1}$ and $\mathbf{v}\in\Delta^{M-1}$ denote the marginal distributions over token positions for the model and the corresponding ExpCoT trace, obtained by normalizing their generation probabilities. The OT distance for response $\mathbf{y}^{i}$ is defined by
\begin{equation}
\label{eq:ot_distance}
\begin{aligned}
W^{i} &= \min_{\mathbf{P}^{i}\in\Pi(\mathbf{u}^{i},\mathbf{v})}
\langle \mathbf{P}^{i},\mathbf{C}^{i}\rangle_F, \\
\Pi(\mathbf{u}^{i},\mathbf{v})
&= \{\mathbf{P}^{i}\!\ge 0 \mid
\mathbf{P}^{i}\mathbbm{1}_{M}=\mathbf{u}^{i},
(\mathbf{P}^{i})^\top\mathbbm{1}_{N^i}=\mathbf{v}\},
\end{aligned}
\end{equation}
where $\langle\cdot,\cdot\rangle_F$ is the Frobenius product,  and $W^{i}$ measures the minimal semantic effort required to transform the model’s reasoning distribution into the ExpCoT distribution. In practice, we approximate the solution of Equation~(\ref{eq:ot_distance}) with the entropic-regularized Sinkhorn–Knopp~\citep{cuturi2013sinkhorn} for efficiency and smoothness. Then, we use $r^{i}_{\mathrm{TsR}}=\exp(-W^{i}/\tau)$ as the reasoning reward and obtain the \emph{time-series reasoning advantage}:
\begin{equation}
\label{eq:semantic_advantage_hat}
\widehat{A}_{\mathrm{TsR}}^{i}
= \dfrac{r^{i}_{\mathrm{TsR}} - \mu_{\mathrm{TsR}}}{\sigma_{\mathrm{TsR}} + \varepsilon},
\quad
\mu_{\mathrm{TsR}}
= \dfrac{1}{G} \sum_{i=1}^{G} r^{i}_{\mathrm{TsR}},
\end{equation}
where $\sigma_{\mathrm{TsR}}^{2}
= \dfrac{1}{G} \sum_{i=1}^{G} \left(r^{i}_{\mathrm{TsR}} - \mu_{\mathrm{TsR}}\right)^{2}$.
By collecting the values across the group $\mathcal{G}$, we obtain
$\widehat{A}_{\mathrm{TsR}} = (\widehat{A}_{\mathrm{TsR}}^{1}, \dots, \widehat{A}_{\mathrm{TsR}}^{G})^\top \in \mathbb{R}^{G}$,
which serves as a relative measure of reasoning quality.

\paragraph{Orthogonal Integration for Policy Optimization.}
A naive combination of task and reasoning rewards risks interference, as ExpCoT guidance may overlap with the primary detection objective under shared ground truth supervision. To avoid this, TimerPO orthogonalizes the time-series grounded advantage with respect to the main advantage, retaining only the complementary part:
\begin{equation}
\label{eq:orthogonalization_hat}
\widehat{A}_{\mathrm{TsR}}^{\perp}=\widehat{A}_{\mathrm{TsR}}-\frac{\langle \widehat{A}_{\mathrm{TsR}},\,\widehat{A}_{\mathrm{main}}\rangle}
     {\|\widehat{A}_{\mathrm{main}}\|_2^{2}+\varepsilon}\,\widehat{A}_{\mathrm{main}}.
\end{equation}
We then compose the final advantage for each response by
\begin{equation}
\label{eq:final_advantage_hat}
A_{\mathrm{final}}^{i}=\widehat{A}_{\mathrm{main}}^{i}+\alpha\,\big(\widehat{A}_{\mathrm{TsR}}^{\perp}\big)^{i},\qquad i=1,\dots,G,
\end{equation}
where $\alpha$ is a hyperparameter controlling the strength of the reasoning refinement. This composite advantage, $A_{\mathrm{final}}^{i}$, then drives the policy update by replacing the standard normalized advantage in the clipped objective function:
\begin{equation}
\footnotesize
\label{eq:grpo_final}
\begin{aligned}
\mathcal{L}(\theta)
&=\frac{1}{G}\sum_{i=1}^{G}\frac{1}{|\mathbf y^{i}|}
\sum_{n=1}^{|\mathbf y^{i}|}
\min\!\Big(
\rho^{i}_{n}A^{i}_{\mathrm{final}},\;
\tilde{A}^{\,i}_{n}
\Big)
-\beta\,\mathrm{KL}\!\big[\pi_\theta\|\pi_{\mathrm{ref}}\big],
\end{aligned}
\end{equation}
where $\rho^{\,i}_{n}$ is the importance ratio for the $n$-th token of response $\mathbf{y}^i$, and
$
\tilde{A}^{\,i}_{n}
=
\mathrm{clip}(\rho^{i}_{n},1-\epsilon,1+\epsilon)\,A^{i}_{\mathrm{final}},
$
with $\epsilon$ and $\beta$ denoting the PPO clipping and KL coefficients, respectively. By operating at the advantage level, TimerPO offers a stable mechanism to instill ExpCoT reasoning, enhancing the model's analytical precision while keeping the primary detection update direction unchanged.

\begin{figure}[t]
    \centering
  \includegraphics[width=1\linewidth]{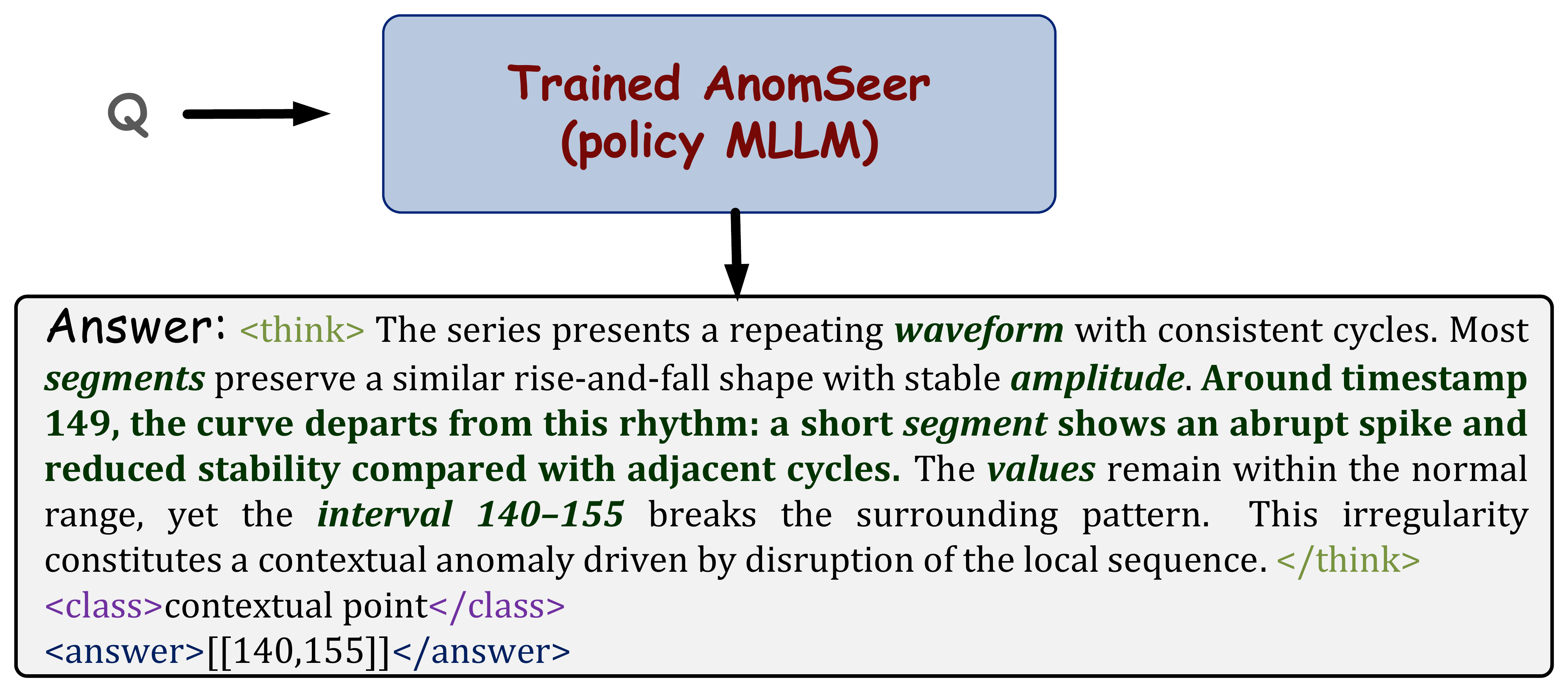}
    \caption{An example of TSAD reasoning by \methodname{} at inference. The model runs fully end-to-end, without relying on ExpCoT traces or classical detectors.}
    \label{fig:inf}
\vspace{-10pt}
\end{figure}

\noindent\textbf{Overall.}
\methodname{} employs the pure RL training strategy to enhance MLLMs without SFT as a cold-start or any modifications to the model architecture.
During training, we first construct ExpCoT using the analytical rigor of traditional TSAD methods, and subsequently refine the model’s policy using orthogonalized time-series reasoning advantages through TimerPO. This simple yet effective integrated design efficiently instills expert knowledge into the pre-trained model within a single reinforcement learning phase. At inference time, \methodname{} operates in a fully end-to-end manner, requiring no external components or incurring any additional token overhead. 
As shown in Figure~\ref{fig:inf}, the trained \methodname{} receives the question and produces outputs that include step-by-step analysis, anomaly type classification, and precise interval localization. Appendix~\ref{app:pseudo} provides the pseudocode for the overall \methodname{} procedure.

\section{Experiments}

\textbf{Benchmarks.}~~
To evaluate the performance and generalization ability of \methodname{}, we consider three diverse TSAD benchmarks:
(\textbf{1}) \emph{AnomLLM}~\citep{zhou2024can}, a synthetic dataset containing frequency, trend, out-of-range and point anomalies\footnote{In {AnomLLM}, contextual \textit{frequency}, \textit{trend}, and \textit{point} anomalies are harder as they require contextual awareness, and \textit{range} anomalies are easier as they show obvious global point deviations.};
(\textbf{2}) \emph{VisualTimeAnomaly}~\citep{xu2025can}, a mixed synthetic–real, image-based benchmark covering a broader spectrum of anomaly types\footnote{In {VisualTimeAnomaly}, \textit{range-wise} anomalies (shapelet, seasonal, and trend) are generally easier, while \textit{point-wise contextual} and \textit{global} anomalies, which  manifest as subtle and dispersed single points, are harder.}; and
(\textbf{3}) \emph{TSB-UAD}~\citep{paparrizos2022tsb, qiu2025tab}, a real-world univariate collection from domains such as ECG and web traffic, with diverse anomaly types, ratios, and sequence lengths.
Training is conducted solely on the synthetic AnomLLM benchmark (3,200 instances), ensuring clean, high-fidelity ExpCoT supervision. Evaluation is then performed on the test sets of AnomLLM, the mixed real-world VisualTimeAnomaly, and TSB-UAD, providing a rigorous test of generalization to diverse, previously unseen anomalies.

\textbf{Baselines.}~~
We compare against both commercial (GPT-4o, GPT-4o-mini, Gemini-2.5-Pro, Gemini-2.5-Flash) and open-source MLLMs (Qwen2.5-VL-72B/32B/7B/3B-Instruct), as well as two representative LLM-based temporal reasoning baselines: \emph{SigLLM} (GPT-3.5-based)~\citep{alnegheimish2024large} and \emph{TimeMaster} (Qwen2.5-VL-3B-based, trained with SFT and GRPO)~\citep{zhang2025timemaster}. 
We further compare against SFT baselines: Qwen2.5-VL-3B-SFT3.2k, fine-tuned on 3,200 instances, and Qwen2.5-VL-3B-SFT32k, fine-tuned on 32,000 instances.

\textbf{Metrics.}~~
We report both anomaly-type classification accuracy and label-based metrics for localization performance, including Affinity-Precision (P), Affinity-Recall (R), and Affinity-F1 (F1), following the definitions in \citet{huet2022local}. These metrics are suitable because LLMs generate discrete anomalous intervals, which can be converted into binary labels rather than continuous scores, and they better capture the temporal consistency of anomaly detection~\citep{zhou2024can, xu2025can}.

\textbf{Hyperparameters.}~~
We build \methodname{} on Qwen2.5-VL-3B/7B-Instruct~\citep{bai2025qwen2}. Following \citet{zhang2025timemaster}, we set the group size $G=5$ and the PPO clipping $\epsilon=0.2$. The reward weights are empirically chosen as $\lambda^{\mathrm{fmt}}=0.1$, $\lambda^{\mathrm{cls}}=0.2$, and $\lambda^{\mathrm{loc}}=0.7$. TimerPO’s reasoning advantage weight is fixed at $\alpha=0.3$.
More experimental details are provided in Appendix~\ref{app:exp}.

\begin{table*}[t]
  \centering
\caption{Performance comparison on the AnomLLM test dataset. Results are reported as the mean and standard deviation over three runs for anomaly classification accuracy (\%) and location detection accuracy metrics (\%): Affinity-Precision (P), Affinity-Recall (R), and Affinity-F1 (F1).}
  \label{tab:main_f1}
  \setlength{\tabcolsep}{5pt}
  \renewcommand{\arraystretch}{1.15}
  \resizebox{\textwidth}{!}{
     \begin{tabular}{@{\,\,\,}l@{\,\,\,\,}l@{\,\,\,\,}l@{\,\,\,}c@{\,\,\,}c@{\,\,\,}c@{\,\,\,}c@{\,\,\,}c@{\,\,\,}c@{\,\,\,}c@{\,\,\,}c@{\,\,\,}c@{\,\,\,}c@{\,\,\,}c@{\,\,\,}c@{\,\,\,}c@{\,\,\,}c@{\,\,\,}}
    \toprule
    \multirow{3}{*}{\textbf{Modality}} &
    \multirow{3}{*}{\textbf{Type}} &
    \multirow{3}{*}{\textbf{Method}} &
    \multicolumn{1}{c}{\textbf{Classification}} &
    \multicolumn{13}{c}{\textbf{Location}} \\
    \cmidrule(lr){4-4} \cmidrule(lr){5-17}
    & & &
    \multirow{2}{*}{\textbf{Accuracy}} &
    \multicolumn{3}{c}{\textbf{Frequency}} &
    \multicolumn{3}{c}{\textbf{Trend}} &
    \multicolumn{3}{c}{\textbf{Range}} &
    \multicolumn{3}{c}{\textbf{Point}} &
    \multirow{2}{*}{\textbf{Avg F1}} \\
    \cmidrule(lr){5-7} \cmidrule(lr){8-10} \cmidrule(lr){11-13} \cmidrule(lr){14-16}
    & & &
    & \textbf{P} & \textbf{R} & \textbf{F1}
    & \textbf{P} & \textbf{R} & \textbf{F1}
    & \textbf{P} & \textbf{R} & \textbf{F1}
    & \textbf{P} & \textbf{R} & \textbf{F1}
    & \\
    \midrule
    \multicolumn{17}{l}{\emph{Commercial}} \\
    Image+Text & Prompting & GPT-4o      & 17.2\(_{\scriptscriptstyle\pm 1.1}\) & 11.1\(_{\scriptscriptstyle\pm 0.2}\) & 10.8\(_{\scriptscriptstyle\pm 0.2}\) & 10.9\(_{\scriptscriptstyle\pm 0.2}\) & 40.3\(_{\scriptscriptstyle\pm 0.4}\) &48.4\(_{\scriptscriptstyle\pm 0.1}\) & 43.5\(_{\scriptscriptstyle\pm 0.2}\) &55.0\(_{\scriptscriptstyle\pm 0.5}\) &61.8\(_{\scriptscriptstyle\pm 0.5}\) &57.0\(_{\scriptscriptstyle\pm 0.5}\)& 51.5\(_{\scriptscriptstyle\pm 0.2}\) & 58.8\(_{\scriptscriptstyle\pm 0.2}\) &53.4\(_{\scriptscriptstyle\pm 0.2}\) & 41.2 \\
    Image+Text & Prompting & GPT-4o-mini    & 17.8\(_{\scriptscriptstyle\pm 1.2}\) &     
    
    10.3\(_{\scriptscriptstyle\pm 0.1}\) & 10.1 \(_{\scriptscriptstyle\pm 0.1}\) &10.2 \(_{\scriptscriptstyle\pm 0.1}\)& 19.4\(_{\scriptscriptstyle\pm 0.2}\) & 29.4 \(_{\scriptscriptstyle\pm 0.2}\) & 23.2 \(_{\scriptscriptstyle\pm 0.2}\) & 48.0 \(_{\scriptscriptstyle\pm 0.1}\) & 58.6 \(_{\scriptscriptstyle\pm 0.1}\) & 51.3 \(_{\scriptscriptstyle\pm 0.1}\) & 51.3 \(_{\scriptscriptstyle\pm 0.1}\) & 58.2 \(_{\scriptscriptstyle\pm 0.1}\) & 52.7 \(_{\scriptscriptstyle\pm 0.1}\) & 34.4 \\
    Image+Text & Prompting & Gemini-2.5-Flash      &  10.0 \(_{\scriptscriptstyle\pm 0.5}\)  &  21.4 \(_{\scriptscriptstyle\pm 0.1}\) & 16.6 \(_{\scriptscriptstyle\pm 0.1}\) & 17.9 \(_{\scriptscriptstyle\pm 0.1}\) 
    & 34.6 \(_{\scriptscriptstyle\pm 0.1}\) & 36.0 \(_{\scriptscriptstyle\pm 0.1}\) & 35.2 \(_{\scriptscriptstyle\pm 0.1}\) & 76.0 \(_{\scriptscriptstyle\pm 0.1}\) & 78.9 \(_{\scriptscriptstyle\pm 0.1}\) & 76.7 \(_{\scriptscriptstyle\pm 0.1}\) & 76.4 \(_{\scriptscriptstyle\pm 0.5}\) & 74.9 \(_{\scriptscriptstyle\pm 0.5}\) &74.9 \(_{\scriptscriptstyle\pm 0.5}\) & 51.2  \\
    Image+Text & Prompting & Gemini-2.5-Pro   &  12.6 \(_{\scriptscriptstyle\pm 0.1}\) &     
    17.4\(_{\scriptscriptstyle\pm 0.1}\) & 22.0\(_{\scriptscriptstyle\pm 0.1}\)& 19.1\(_{\scriptscriptstyle\pm 0.3}\)  & 58.8 \(_{\scriptscriptstyle\pm 0.5}\) & 60.0 \(_{\scriptscriptstyle\pm 0.1}\) & 59.0 \(_{\scriptscriptstyle\pm 0.2}\) & 79.4 \(_{\scriptscriptstyle\pm 0.1}\) & 83.2 \(_{\scriptscriptstyle\pm 0.5}\) & 81.3 \(_{\scriptscriptstyle\pm 0.4}\) & 76.1 \(_{\scriptscriptstyle\pm 0.5}\) & 74.4 \(_{\scriptscriptstyle\pm 0.4}\) & 74.5 \(_{\scriptscriptstyle\pm 0.5}\) & 58.5 \\
    Numerical+Text & Prompting & SigLLM (GPT-3.5)  & \textbackslash{} & 16.9 \(_{\scriptscriptstyle\pm 0.5}\) & 14.9 \(_{\scriptscriptstyle\pm 0.5}\) & 15.8 \(_{\scriptscriptstyle\pm 0.5}\) & 20.3 \(_{\scriptscriptstyle\pm 0.5}\) & 20.5 \(_{\scriptscriptstyle\pm 0.5}\) & 19.6 \(_{\scriptscriptstyle\pm 0.1}\) & 67.8 \(_{\scriptscriptstyle\pm 0.1}\) & 67.5 \(_{\scriptscriptstyle\pm 0.1}\) & 67.7 \(_{\scriptscriptstyle\pm 0.1}\) & 34.4 \(_{\scriptscriptstyle\pm 0.1}\) &38.6 \(_{\scriptscriptstyle\pm 0.1}\) & 36.2 \(_{\scriptscriptstyle\pm 0.1}\) & 34.8 \\

    \midrule
    \multicolumn{17}{l}{\emph{Open-source}} \\
    Image+Text & Prompting & Qwen2.5-VL-72B-Instruct & 14.6\(_{\scriptscriptstyle\pm 0.5}\) & 40.2\(_{\scriptscriptstyle\pm 0.1}\) & 28.3\(_{\scriptscriptstyle\pm 0.1}\)& 31.4\(_{\scriptscriptstyle\pm 0.1}\) & 30.8\(_{\scriptscriptstyle\pm 0.2}\) & 33.8\(_{\scriptscriptstyle\pm 0.1}\) & 32.1\(_{\scriptscriptstyle\pm 0.7}\) & 76.8\(_{\scriptscriptstyle\pm 0.4}\) & 73.9\(_{\scriptscriptstyle\pm 0.1}\) & 74.6\(_{\scriptscriptstyle\pm 0.1}\)&63.2\(_{\scriptscriptstyle\pm 0.5}\) & 64.6\(_{\scriptscriptstyle\pm 0.3}\) &62.7\(_{\scriptscriptstyle\pm 0.1}\) & 50.2 \\
        Image+Text & Prompting &Qwen2.5-VL-32B-Instruct   & 10.2\(_{\scriptscriptstyle\pm 0.5}\) & 19.3\(_{\scriptscriptstyle\pm 0.4}\) & 20.2\(_{\scriptscriptstyle\pm 0.5}\) & 18.9\(_{\scriptscriptstyle\pm 0.2}\) & 34.3\(_{\scriptscriptstyle\pm 0.5}\) & 37.2\(_{\scriptscriptstyle\pm 0.2}\) & 35.5\(_{\scriptscriptstyle\pm 0.5}\) & 72.0\(_{\scriptscriptstyle\pm 0.5}\) & 70.3\(_{\scriptscriptstyle\pm 0.5}\)  &  70.7\(_{\scriptscriptstyle\pm 0.5}\) & 55.9\(_{\scriptscriptstyle\pm 0.5}\) & 59.3\(_{\scriptscriptstyle\pm 0.5}\) & 56.7\(_{\scriptscriptstyle\pm 0.5}\) & 45.5 \\
    Image+Text & Prompting & Qwen2.5-VL-7B-Instruct   & 25.3 \(_{\scriptscriptstyle\pm 0.2}\) & 18.5 \(_{\scriptscriptstyle\pm 0.4}\) & 16.4 \(_{\scriptscriptstyle\pm 0.1}\) & 16.8   \(_{\scriptscriptstyle\pm 0.1}\) & 52.7 \(_{\scriptscriptstyle\pm 0.1}\) & 53.8 \(_{\scriptscriptstyle\pm 0.2}\) & 53.1 \(_{\scriptscriptstyle\pm 0.3}\) & 48.6 \(_{\scriptscriptstyle\pm 0.4}\) & 45.4 \(_{\scriptscriptstyle\pm 0.1}\) & 46.4 \(_{\scriptscriptstyle\pm 0.1}\) & 61.9 \(_{\scriptscriptstyle\pm 0.1}\) & 58.6 \(_{\scriptscriptstyle\pm 0.1}\) & 59.0 \(_{\scriptscriptstyle\pm 0.1}\) & 43.8 \\
    Image+Text & Prompting & Qwen2.5-VL-3B-Instruct   & 11.4 \(_{\scriptscriptstyle\pm 0.1}\) & 7.1 \(_{\scriptscriptstyle\pm 0.1}\) & 9.3 \(_{\scriptscriptstyle\pm 0.2}\) & 7.9 \(_{\scriptscriptstyle\pm 0.4}\) & 17.4\(_{\scriptscriptstyle\pm 0.5}\) & 22.0 \(_{\scriptscriptstyle\pm 0.5}\) & 19.1 \(_{\scriptscriptstyle\pm 0.1}\) & 29.7 \(_{\scriptscriptstyle\pm 0.1}\) & 31.3 \(_{\scriptscriptstyle\pm 0.1}\) & 29.7 \(_{\scriptscriptstyle\pm 0.1}\) & 34.3 \(_{\scriptscriptstyle\pm 0.5}\) & 40.6 \(_{\scriptscriptstyle\pm 0.5}\) & 36.2 \(_{\scriptscriptstyle\pm 0.1}\) & 23.2 \\
   
    Image+Text & Training  & Qwen2.5-VL-3B-SFT3.2K  & 29.7 \(_{\scriptscriptstyle\pm 0.1}\) & 19.0 \(_{\scriptscriptstyle\pm 0.1}\) & 24.4 \(_{\scriptscriptstyle\pm 0.5}\) & 19.7 \(_{\scriptscriptstyle\pm 0.1}\) & 29.7 \(_{\scriptscriptstyle\pm 0.2}\) & 34.8 \(_{\scriptscriptstyle\pm 0.1}\) & 30.2 \(_{\scriptscriptstyle\pm 0.2}\) & 40.1 \(_{\scriptscriptstyle\pm 0.1}\) & 48.1 \(_{\scriptscriptstyle\pm  0.4}\) & 40.8 \(_{\scriptscriptstyle\pm 0.1}\) & 40.4 \(_{\scriptscriptstyle\pm 0.5}\) & 49.3 \(_{\scriptscriptstyle\pm 0.1}\) & 40.6 \(_{\scriptscriptstyle\pm 0.2}\) & 32.8 \\
    Image+Text & Training  & Qwen2.5-VL-3B-SFT32K  & 35.6 \(_{\scriptscriptstyle\pm 0.5}\) &  12.0 \(_{\scriptscriptstyle\pm 0.5}\) & 14.3 \(_{\scriptscriptstyle\pm 0.1}\) & 12.4 \(_{\scriptscriptstyle\pm 0.1}\) & 57.6 \(_{\scriptscriptstyle\pm 0.5}\) & 57.5 \(_{\scriptscriptstyle\pm 0.2}\) & 57.4 \(_{\scriptscriptstyle\pm 0.5}\) & 40.4 \(_{\scriptscriptstyle\pm 0.1}\) & 52.1 \(_{\scriptscriptstyle\pm 0.1}\) & 41.3 \(_{\scriptscriptstyle\pm 0.2}\) & 44.3 \(_{\scriptscriptstyle\pm 0.5}\) & 58.3 \(_{\scriptscriptstyle\pm 0.2}\) & 46.3 \(_{\scriptscriptstyle\pm 0.5}\) & 39.4 \\
     Image+Text & Training  & TimeMaster-3B & 57.9 \(_{\scriptscriptstyle\pm 0.6}\) & 57.3 \(_{\scriptscriptstyle\pm 0.5}\)  & 50.3 \(_{\scriptscriptstyle\pm 0.1}\)  &51.4 \(_{\scriptscriptstyle\pm 0.2}\)  & 76.0 \(_{\scriptscriptstyle\pm 0.5}\) & 77.3 \(_{\scriptscriptstyle\pm 0.1}\)  & 76.6 \(_{\scriptscriptstyle\pm 0.5}\)  & 77.8 \(_{\scriptscriptstyle\pm 0.5}\) &83.5\(_{\scriptscriptstyle\pm 0.1}\) & 80.1 \(_{\scriptscriptstyle\pm 0.1}\) & 77.7 \(_{\scriptscriptstyle\pm 0.5}\) &82.1 \(_{\scriptscriptstyle\pm 0.1}\) & 79.6 \(_{\scriptscriptstyle\pm 0.5}\) & 71.9 \\
    \rowcolor{gray!15} Image+Text & Training & \methodname{}-3B (Ours) & 62.8\(_{\scriptscriptstyle\pm 0.5}\)  & 63.7\(_{\scriptscriptstyle\pm 0.5}\)  & 58.4\(_{\scriptscriptstyle\pm 0.5}\) & 58.9\(_{\scriptscriptstyle\pm 0.5}\) & 84.2\(_{\scriptscriptstyle\pm 0.2}\) & 85.9\(_{\scriptscriptstyle\pm 0.1}\) & 84.9\(_{\scriptscriptstyle\pm 0.1}\) & 83.3\(_{\scriptscriptstyle\pm 0.3}\) & 89.2\(_{\scriptscriptstyle\pm 0.1}\) & 85.6\(_{\scriptscriptstyle\pm 0.1}\) & 86.0\(_{\scriptscriptstyle\pm 0.1}\) & 90.3\(_{\scriptscriptstyle\pm 0.1}\) & 87.8\(_{\scriptscriptstyle\pm 0.1}\) & 79.3 \\
    \rowcolor{gray!15} Image+Text & Training & \methodname{}-7B (Ours) &  65.0\(_{\scriptscriptstyle\pm 0.5}\) &  68.3\(_{\scriptscriptstyle\pm 0.5}\) &  59.4\(_{\scriptscriptstyle\pm 0.2}\) &  60.8\(_{\scriptscriptstyle\pm 0.2}\) &  86.6\(_{\scriptscriptstyle\pm 0.5}\) &  89.0\(_{\scriptscriptstyle\pm 0.5}\)    &  87.7    \(_{\scriptscriptstyle\pm 0.5}\) & 91.6    \(_{\scriptscriptstyle\pm 0.1}\)  & 97.8    \(_{\scriptscriptstyle\pm 0.4}\)  & 94.3    \(_{\scriptscriptstyle\pm 0.1}\)  & 93.4    \(_{\scriptscriptstyle\pm 0.1}\)  & 96.9    \(_{\scriptscriptstyle\pm 0.5}\)  & 94.9    \(_{\scriptscriptstyle\pm 0.1}\)  & 84.4 \\
    \bottomrule
  \end{tabular}
  }

\end{table*}

\begin{table}[htbp]
\centering
\caption{Ablation study on different components of \methodname{}-3B using Affinity F1 score (\%).}
\label{tab:ablation}

\resizebox{\columnwidth}{!}{
\begin{tabular}{ccc|cccc}
    \toprule
    \multicolumn{3}{c|}{\textbf{Components}} & \multicolumn{4}{c}{\textbf{Anomaly Scenarios}} \\
    \midrule
    ExpCoT & $\widehat{A}_{\mathrm{TsR}}^{\perp}$ & Orth & Frequency & Trend & Range & Point \\
    \midrule
    \xmark & \checkmark & \checkmark & 49.8 & 79.5 & 84.4 & 86.1 \\
    \checkmark & \checkmark & \xmark & 53.5 & 81.1 & 83.5 & 85.4 \\
    \xmark & \xmark & \xmark & 50.4 & 77.8 & 81.8 & 80.6 \\
    \checkmark & \checkmark & \checkmark & \textbf{58.9} & \textbf{84.9} & \textbf{85.6} & \textbf{87.8} \\
    \bottomrule
\end{tabular}
}

\end{table}

\subsection{Main Results}
As shown in Table~\ref{tab:main_f1}, \methodname{} consistently achieves state-of-the-art results across all anomaly detection tasks on the {AnomLLM} benchmark. Remarkably, even at a lightweight 3B scale, our model substantially outperforms much larger and more resource-intensive MLLMs such as GPT-4o and Gemini-2.5-Pro in both anomaly type classification and Affinity-F1 metrics, and its performance further improves with the 7B variant.
We also observe that simply increasing the amount of SFT data yields only marginal gains, even with $10\times$ more SFT data (32k instances), performance still falls short of \methodname{}. One possible reason is that SFT emphasizes only positive reasoning paths while neglecting negative ones, leading the model to develop only a shallow understanding rather than genuinely learning.
Notably, for numerically subtle anomalies such as frequency shifts, \methodname{} maintains a clear advantage, whereas GRPO-trained MLLMs like TimeMaster continue to lag behind. This result suggests that globally verifiable RL objectives alone are insufficient for modeling fine-grained temporal variations, whereas our \methodname{} explicitly encourages fine-grained temporal reasoning that leads to more accurate anomaly detection.

\begin{figure}[t]
    \centering
    \includegraphics[width=0.95\columnwidth]{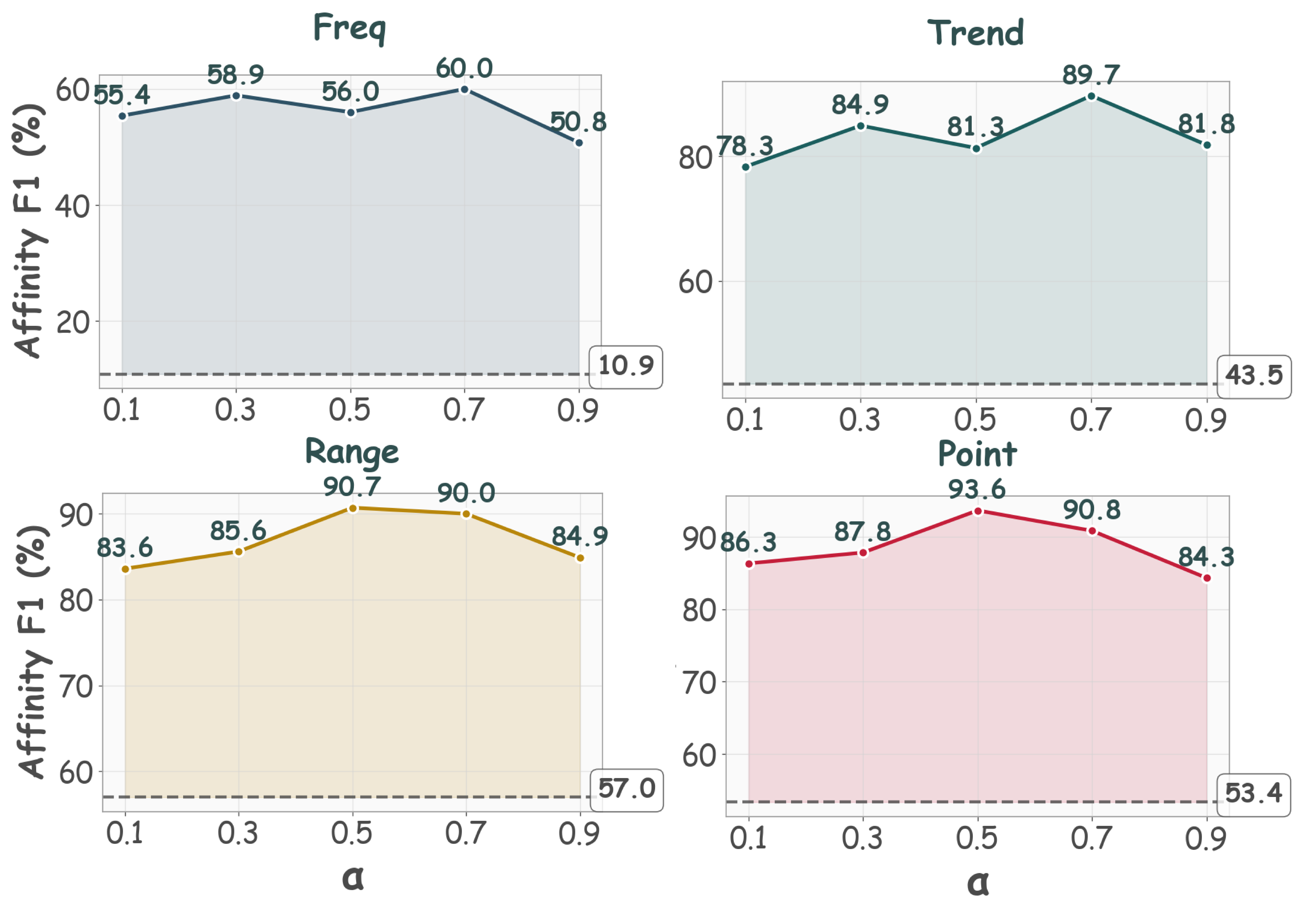}
    \caption{Hyperparameter sensitivity analysis on $\alpha$, comparing our method with the GPT-4o baseline (grey dashed line).}
    \label{fig:sensitivity}

\end{figure}

\begin{figure*}[t]
\centering
\includegraphics[width=0.95\textwidth]{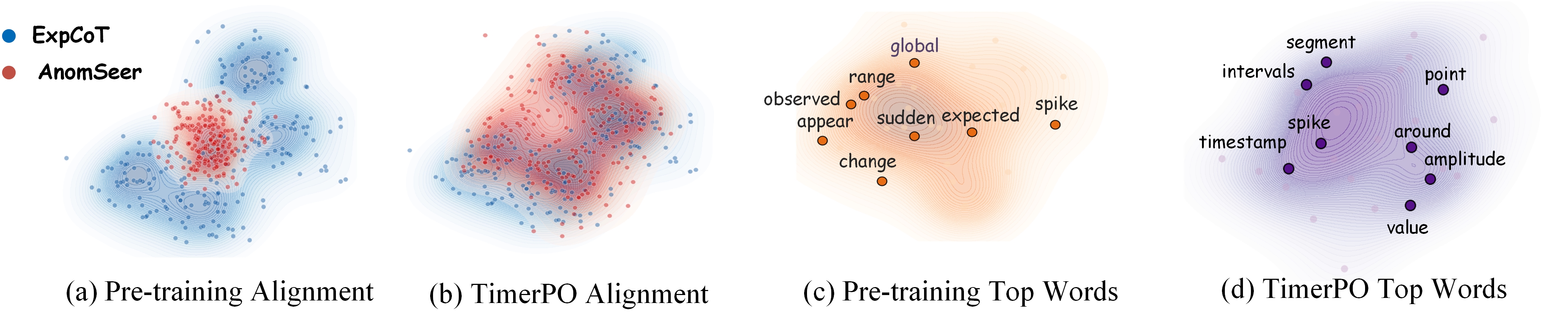}
\vspace{-0.1in}
\caption{Comparison of distribution alignment between ExpCoT (blue) and \methodname{} (red) outputs, as well as token usage before and after applying TimerPO. }
\label{fig:alignment}

\end{figure*}

\subsection{Ablation Study and Hyperparameter Analysis}
We next conduct a detailed ablation study together with a hyperparameter sensitivity analysis. Table~\ref{tab:ablation} provides several key takeaways. First, we replace ExpCoT with CoT generated by GPT-4o, which leads to a marked degradation, particularly on challenging frequency anomalies. This demonstrates that generic CoT supervision imparts mere surface-level fluency rather than  in-depth temporal reasoning. It further highlights a crucial insight: the analytical rigor of classical methods is not obsolete, but rather a valuable resource for shaping the next generation of truly capable time-series MLLMs. Second, removing the orthogonalization mechanism causes a moderate drop in performance, underscoring its crucial role in mitigating spurious correlations between reasoning quality and task success. Third, 
eliminating all components reduces the method to a vanilla GRPO setup and yields the worst average performance, confirming that outcome-based rewards alone are insufficient to foster the fine-grained anomaly detection skills required for complex TSAD.


Figure~\ref{fig:sensitivity} presents the effect of varying the temporal reasoning weight $\alpha$ in our TimerPO objective. Across all anomaly types, \methodname{}  maintains a substantial margin over the GPT-4o baseline (grey dashed line), showing that even under suboptimal $\alpha$ values, the integration of structured temporal reasoning signals offers clear benefits. The model remains relatively robust within the range $\alpha \in [0.3, 0.7]$, where performance is stable and near-optimal for frequency, trend, range, and point anomalies alike. This highlights the importance of balancing outcome-level and reasoning-level rewards: too small a weight diminishes the impact of explicit reasoning supervision, while too large a weight can overshadow task-level alignment, leading to slight degradation. In practice, $\alpha=0.3$ works well as a default, though dataset-specific tuning may yield more gains.

\subsection{Effect of TimerPO on Reasoning Pattern}

To show that \methodname{} enables time-series MLLMs reasoning grounded in fine-grained statistics, we analyze the effect of TimerPO on distributional alignment and linguistic usage before and after RL training, as shown in Figure~\ref{fig:alignment}.
Panels (a)-(b) illustrate that, prior to TimerPO, ExpCoT (blue) and \methodname{} outputs (red) occupy noticeably divergent regions in the representation space, with the latter exhibiting a relatively narrow distribution. This mismatch highlights that the model’s reasoning is overly global and lacks diversity.
A similar trend is observed in token usage. In the pre-training stage (c), top words are generic and coarse-grained (e.g., global, sudden, change), reflecting surface-level anomaly descriptions. After TimerPO (d), the vocabulary shifts toward finer-grained and temporally grounded tokens (e.g., timestamp, intervals, amplitude), which better capture structured reasoning over time.
Therefore, these results demonstrate that TimerPO not only improves distributional alignment with expert reasoning but also enriches the semantic granularity of reasoning traces, moving from broad anomaly descriptors to precise temporal markers. We also compare GRPO and our TimerPO-trained models in Appendix~\ref{app:res1}, which further confirms the effectiveness of our method in enhancing temporal reasoning.

\subsection{Generalization Performance}
\begin{figure}[t]
    \centering
    \includegraphics[width=0.85\columnwidth]{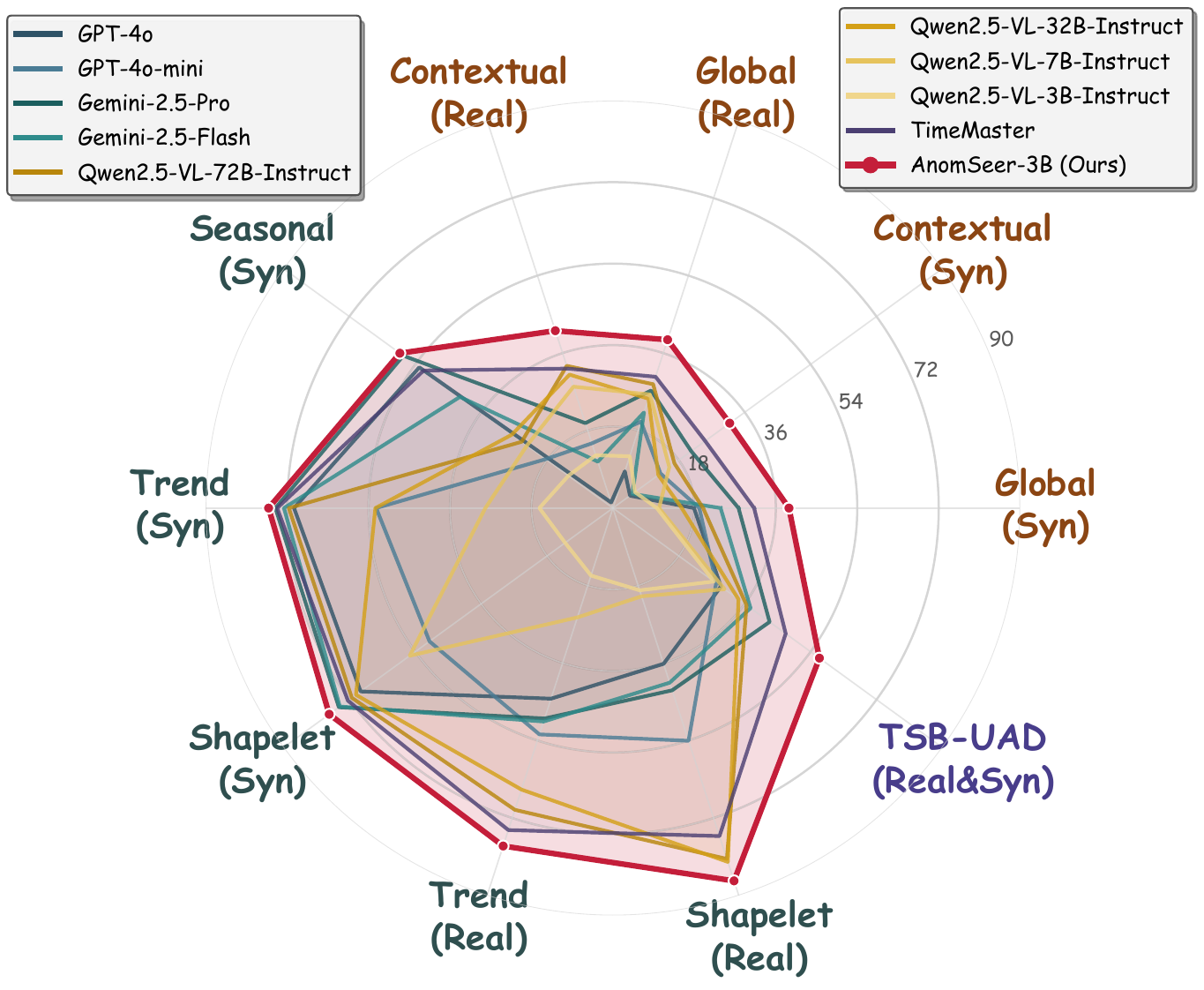}
    \caption{Comparison of model generalization performance (Affinity F1\%) across 
    \textcolor[HTML]{8B4513}{point-wise} tasks, 
    \textcolor[HTML]{2F4F4F}{range-wise} tasks, 
    and the real-world 
    \textcolor[HTML]{483D8B}{TSB-UAD} benchmark.}
    \label{fig:dg}
\vspace{-15pt}
\end{figure}

At last, we evaluate the generalization ability of \methodname{}.
We test the model (trained on the synthetic AnomLLM) on two distinct and more challenging benchmarks: VisualTimeAnomaly (a hybrid synthetic-real dataset with richer anomaly types) and TSB-UAD (a real-world univariate collection).
Importantly, shapelet anomalies represent a completely new category absent during training. Despite this, as shown in Figure~\ref{fig:dg}, our method demonstrates strong accuracy on such cases. This ability to detect and explain shapelet anomalies shows that the model is not restricted to pattern memorization, but can generalize to qualitatively novel anomaly behaviors. Moreover, on point-wise contextual anomalies, a notably harder task that requires fine-grained discrimination, \methodname{} delivers clear gains over baseline MLLMs, underscoring its ability to move beyond surface-level visual cues.
Finally, on the TSB-UAD collection of real-world datasets, which spans diverse  domains, \methodname{} sustains its advantage and confirms that the improvements extend beyond synthetic benchmarks to practical anomaly detection scenarios. Overall, these results verify that our approach achieves not only high in-domain accuracy but also robust generalization to unseen and real-world anomalies.

\section{Conclusions and Limitations}
In this paper, we introduced \methodname{}, an RL post-training method that enables multimodal LLMs to detect and reason about time-series anomalies in a fine-grained and accurate manner. By grounding MLLMs’ reasoning in the fine-grained, multi-dimensional evidence of classical TSAD, \methodname{} attains state-of-the-art performance across diverse benchmarks. Beyond surpassing strong baselines such as GPT-4o in detection accuracy and localization, it delivers verifiable, detailed time-series explanations, elevating MLLMs from coarse visual heuristics to principled, testable analysis. 
Nevertheless, \methodname{} was developed primarily on univariate time-series data in TSAD, and extending it to more complex multivariate scenarios remains an open direction. A potential solution is to reframe each variable as an image-like subrepresentation and then reason over its joint structure, enabling the model to capture both localized temporal patterns and cross-variable dependencies in a coherent manner. Another direction may be to explore how to incorporate external knowledge to better account for real-world events that drive anomaly dynamics.

\section*{Acknowledgements}
This work is supported in part by the National Key R\&D Program of China (2024YFF0907701) and the Ministry of Education, Singapore, under its Academic Research Fund Tier 1 (RG101/24). Xu Guo thanks the support from Wallenberg-NTU Presidential Postdoctoral Fellowship.

We are grateful to Suyu Liu for sharing his expertise in optimal transport and multi-objective optimization, which significantly strengthened this work.

\section*{Impact Statement}
This paper presents work whose goal is to advance the field of Machine
Learning. There are many potential societal consequences of our work, none
which we feel must be specifically highlighted here.

\bibliography{icml2026}
\bibliographystyle{icml2026}

\newpage
\appendix
\onecolumn


\section{Pseudo Code}\label{app:pseudo}
The training pipeline of \methodname{} is provided as follows:
\begin{algorithm}[h]
\caption{Training Time-Series MLLMs with \methodname{}}
\label{alg:anomseer}
\begin{algorithmic}[1]
\STATE {\bfseries Require:} Initial policy $\pi_{\theta_{\text{old}}}$, task distribution $p(\textbf{X})$, discount factor $\gamma$, clipping parameter $\epsilon$, KL penalty $\beta$, group size $G$, ExpCoT generator, TimerPO hyperparameter $\alpha$
\FOR{each training iteration}
    \STATE Update old policy: $\theta_{\text{old}} \leftarrow \theta$
    \STATE \small{\color{gray}{// Data preparation phase}}
    \STATE Sample time-series $\textbf{X} \sim p(\textbf{X})$ and render visualization $I$
    \STATE Generate expert chain-of-thought $\textbf{y}^\star \leftarrow \text{ExpCoT}(\textbf{X})$
    \STATE Construct input $(\textbf{I}, \textbf{c})$

    \STATE \small{\color{gray}{// Advantage computation}}
    \STATE Sample group of responses $\mathcal{G} = \{\textbf{y}^i \sim \pi_{\theta_{\text{old}}}(\cdot|\textbf{I},\textbf{c})\}_{i=1}^G$
    \FOR{each $\textbf{y}^i \in \mathcal{G}$}
        \STATE Compute outcome reward: $r^{i} = \lambda^{\mathrm{fmt}} r^{\mathrm{fmt},\,i} \;+\; \lambda^{\mathrm{cls}} r^{\mathrm{cls},\,i} \;+\; \lambda^{\mathrm{loc}} r^{\mathrm{loc},\,i}$
        \STATE Normalize to obtain outcome-aware advantage $\widehat{A}^i_{\mathrm{main}}$ via Eq.~(\ref{eq:main_advantage_hat})
        \STATE Compute semantic OT distance $W^i$ between $\textbf{y}^i$ and $\textbf{y}^\star$  via Eq.~(\ref{eq:ot_distance})
        \STATE Derive reasoning reward $r^{{\rm TsR}}_i = \exp(-W^i/\tau)$ and normalize to $\widehat{A}^i_{\mathrm{TsR}}$  via Eq.~(\ref{eq:semantic_advantage_hat})
    \ENDFOR

    \STATE \small{\color{gray}{// Orthogonal integration of advantages}}
    \STATE Compute orthogonalized reasoning advantage: 
    $$\widehat{A}_{\mathrm{TsR}}^{\perp}=\widehat{A}_{\mathrm{TsR}}-\frac{\langle \widehat{A}_{\mathrm{TsR}},\,\widehat{A}_{\mathrm{main}}\rangle}
     {\|\widehat{A}_{\mathrm{main}}\|_2^{2}+\varepsilon}\,\widehat{A}_{\mathrm{main}}$$
    \STATE Final advantage: $A_{\mathrm{final}}^{i}=\widehat{A}_{\mathrm{main}}^{i}+\alpha\,\big(\widehat{A}_{\mathrm{TsR}}^{\perp}\big)^{i}$

    \STATE \small{\color{gray}{// Policy update}}
    \STATE Update $\theta$ by maximizing the TimerPO objective:
    $$\mathcal{L}(\theta)=\frac{1}{G}\sum_{i=1}^{G}\frac{1}{|\mathbf{y}^{i}|}
\sum_{n=1}^{|\mathbf{y}^{i}|}\min\!\Big(\rho^{i}_{n}\,A_{\mathrm{final}}^{i},\;
\mathrm{clip}(\rho^{i}_{n},\,1-\epsilon,\,1+\epsilon)\,A_{\mathrm{final}}^{i}
\Big)-\beta\,\mathrm{KL}\!\big[\pi_\theta\,\|\,\pi_{\mathrm{ref}}\big],$$
\ENDFOR
\end{algorithmic}
\end{algorithm}

{
\begin{algorithm}[h]
\caption{Inference with \methodname{}}
\label{alg:anomseer_inference}
\begin{algorithmic}[1]
\STATE \textbf{Require:} Trained policy $\pi_\theta$, input time series $\mathbf{X}$, instruction prompt $\textbf{c}$
\STATE Render visualization: $\textbf{I} \leftarrow \mathcal{R}(\mathbf{X})$
\STATE Construct model input: $(\mathbf{I},\mathbf{c})$
\STATE \small{\color{gray}{// Forward inference}}
\STATE Generate model response: $\mathbf{y} \sim \pi_\theta(\cdot \mid \mathbf{I},\mathbf{c})$
\STATE Get output $\mathbf{y}$ including anomaly type, location, and reasoning
\STATE \textbf{return} anomaly prediction results
\end{algorithmic}
\end{algorithm}
}

\section{More Details of \methodname{}}

\subsection{Structured Output for Reasoning.} 
\begin{figure*}[h]
\centering
\resizebox{1\textwidth}{!}{
\begin{tcolorbox}[
  colback=gray!5!white,
  colframe=black!75!black,
  title= Prompt for Anomaly Detection (Image Input),
  boxrule=0.3mm,
  width=\textwidth,
  arc=3mm,
  auto outer arc=true,
  breakable,
  enhanced,
  listing only,
  listing engine=listings,
  listing options={basicstyle=\ttfamily\small, columns=fullflexible, keepspaces=true}
]

    \textbf{Input Image}: 
    \begin{center}
        \includegraphics[width=1\textwidth]{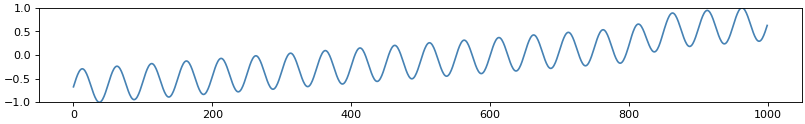}
    \end{center}

    \textbf{Prompt}: 
You are a time series analysis expert. A time series plot of length 1000 is provided.
Your task is to identify anomalous intervals along the x-axis.

The plot shows temporal patterns and typical value ranges based on both the x and y axes and may
exhibit one type of anomaly from the set ["contextual point", "global point", "seasonal", "trend", "shapelet"].
Carefully examine the image to detect intervals that deviate from the expected behavior and
infer the anomaly type.

Begin your detailed reasoning inside a <think>...</think> block.
In this block, analyze the plot, identify suspicious intervals, and explain why they are considered anomalies and the type.

After reasoning, summarize your findings in the following format inside an <answer>...</answer> block:
Each anomaly should be written as a [start, end] pair using x-axis positions.
Then choose the type inside <class>...</class>  from the set ["contextual point", "global point", "seasonal", "trend", "shapelet"].

Examples:
<answer>
[[120, 150], [320, 350]]
</answer>
<class>trend</class>
or
<answer>
[[120, 150]]
</answer>
<class>global point</class>
If no anomalies are found, return:
<answer>[]</answer>
<class>normal</class>
\end{tcolorbox}
}
\vspace{-0.1in}
\caption{Prompt definition for time-series anomaly detection}
\label{fig:tsadprompt}
\end{figure*}

A key objective of \methodname{} is to elicit \emph{textual reasoning} that illuminates the model's analysis process. To achieve this, we enforce a structured output format to decouple the reasoning steps from the final prediction. The model is prompted to first articulate its analytical process within \textcolor{deepgreen}{<think>} \textcolor{deepgreen}{</think>} tags, provide the predicted anomaly category (e.g., trend, global, contextual) within \textcolor{deeppurple}{<class>} \textcolor{deeppurple}{</class>} tags, and present the specific anomalous interval(s) within \textcolor{deepblue}{<answer>} \textcolor{deepblue}{</answer>} tags. This structured prompting strategy bridges low-level visual cues with high-level, human-interpretable reasoning in a unified framework. To illustrate this design, we present our full TSAD prompt in Fig.~\ref{fig:tsadprompt}.

\subsection{Details on ExpCoT.}\label{app:details_expcot}
\begin{figure}[h]
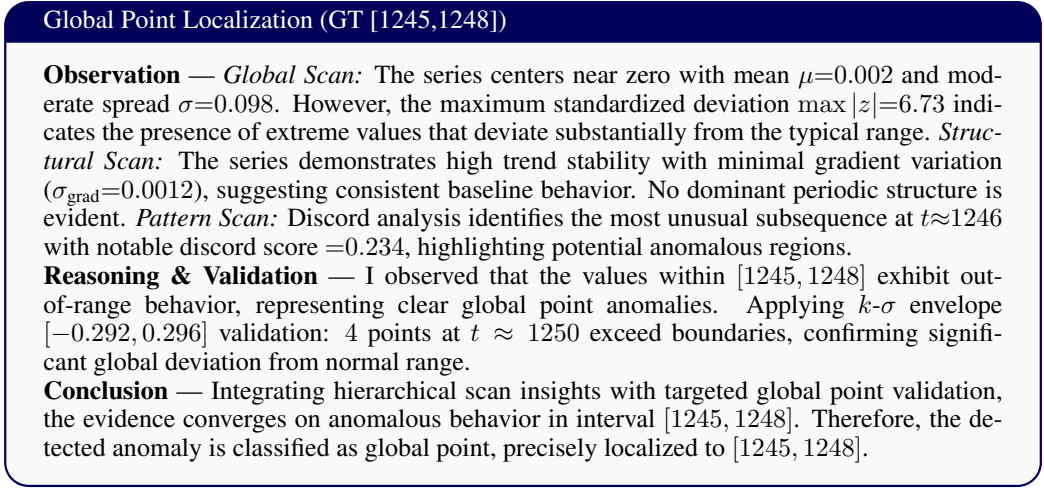

    \centering
    \resizebox{1\linewidth}{!}{
    \begin{tcolorbox}[
      colback=gray!5!white,
      colframe=blue!35!black,
      title={Global Point Localization (GT [1245,1248])},
      boxrule=0.3mm,
      width=\textwidth,
      arc=3mm,
      auto outer arc=true,
      breakable,
      enhanced,
      listing only,
      listing engine=listings,
      listing options={basicstyle=\ttfamily\small, columns=fullflexible, keepspaces=true}
    ]
\textbf{Observation} —  
\emph{Global Scan:} The series centers near zero with mean $\mu{=}0.002$ and moderate spread $\sigma{=}0.098$. However, the maximum standardized deviation $\max|z|{=}6.73$ indicates the presence of extreme values that deviate substantially from the typical range.
\emph{Structural Scan:} The series demonstrates high trend stability with minimal gradient variation ($\sigma_{\text{grad}}{=}0.0012$), suggesting consistent baseline behavior. No dominant periodic structure is evident.
\emph{Pattern Scan:} Discord analysis identifies the most unusual subsequence at $t{\approx}1246$ with notable discord score ${=}0.234$, highlighting potential anomalous regions.

\textbf{Reasoning \& Validation} — I observed that the values within $[1245, 1248]$ exhibit out-of-range behavior, representing clear global point anomalies. Applying $k$-$\sigma$ envelope $[-0.292, 0.296]$ validation: $4$ points  at $t \approx 1250$  exceed boundaries, confirming significant global deviation from normal range.

\textbf{Conclusion} — Integrating hierarchical scan insights with targeted global point validation, the evidence converges on anomalous behavior in interval $[1245, 1248]$. Therefore, the detected anomaly is classified as global point, precisely localized to $[1245, 1248]$.
    \end{tcolorbox}
    }
    \caption{ExpCoT reasoning trace for global point (out-of-range) anomaly detection.}
    \label{fig:expcot_global}
\end{figure}

\begin{figure}[h]
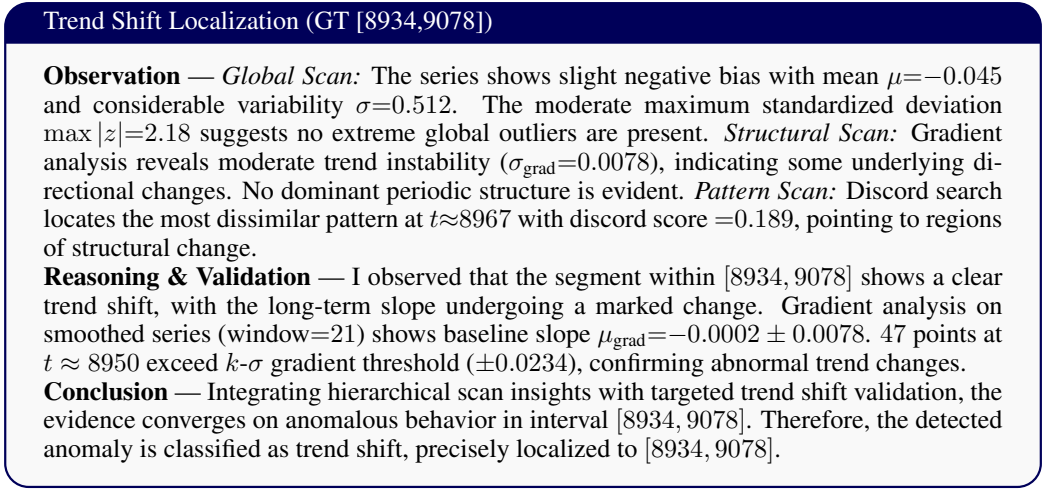

    \centering
    \resizebox{1\linewidth}{!}{
    \begin{tcolorbox}[
      colback=gray!5!white,
      colframe=blue!35!black,
      title={Trend Shift Localization (GT [8934,9078])},
      boxrule=0.3mm,
      width=\textwidth,
      arc=3mm,
      auto outer arc=true,
      breakable,
      enhanced,
      listing only,
      listing engine=listings,
      listing options={basicstyle=\ttfamily\small, columns=fullflexible, keepspaces=true}
    ]
\textbf{Observation} —  
\emph{Global Scan:} The series shows slight negative bias with mean $\mu{=}{-0.045}$ and considerable variability $\sigma{=}0.512$. The moderate maximum standardized deviation $\max|z|{=}2.18$ suggests no extreme global outliers are present.
\emph{Structural Scan:} Gradient analysis reveals moderate trend instability ($\sigma_{\text{grad}}{=}0.0078$), indicating some underlying directional changes. No dominant periodic structure is evident.
\emph{Pattern Scan:} Discord search locates the most dissimilar pattern at $t{\approx}8967$ with discord score ${=}0.189$, pointing to regions of structural change.

\textbf{Reasoning \& Validation} — I observed that the segment within $[8934, 9078]$ shows a clear trend shift, with the long-term slope undergoing a marked change. Gradient analysis on smoothed series (window${=}21$) shows baseline slope $\mu_{\text{grad}}{=}{-0.0002} \pm 0.0078$. $47$ points at $t \approx 8950$ exceed $k$-$\sigma$ gradient threshold ($\pm 0.0234$), confirming abnormal trend changes.

\textbf{Conclusion} — Integrating hierarchical scan insights with targeted trend shift validation, the evidence converges on anomalous behavior in interval $[8934, 9078]$. Therefore, the detected anomaly is classified as trend shift, precisely localized to $[8934, 9078]$.
    \end{tcolorbox}
    }
    \caption{ExpCoT reasoning trace for trend shift anomaly detection.}
    \label{fig:expcot_trend}
\end{figure}

\begin{figure}[h]
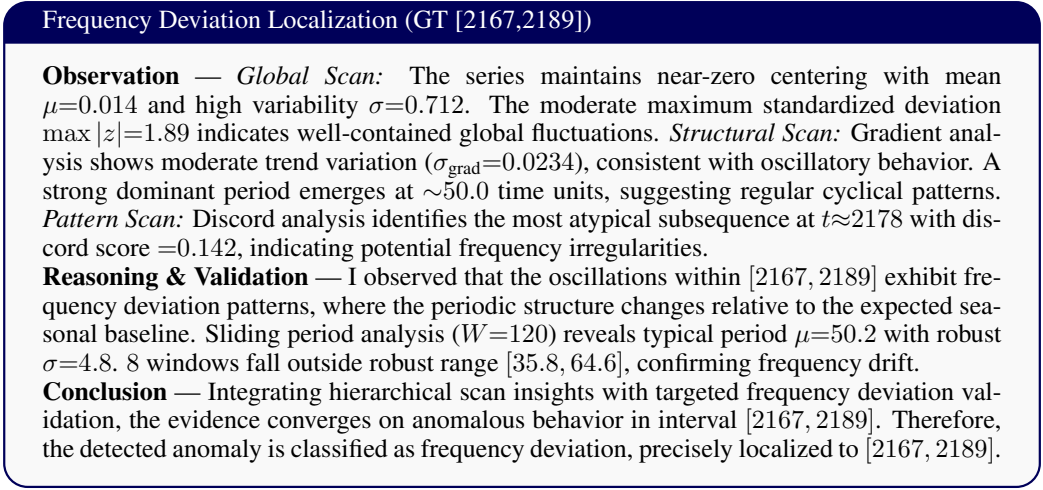

    \centering
    \resizebox{1\linewidth}{!}{
    \begin{tcolorbox}[
      colback=gray!5!white,
      colframe=blue!35!black,
      title={Frequency Deviation Localization (GT [2167,2189])},
      boxrule=0.3mm,
      width=\textwidth,
      arc=3mm,
      auto outer arc=true,
      breakable,
      enhanced,
      listing only,
      listing engine=listings,
      listing options={basicstyle=\ttfamily\small, columns=fullflexible, keepspaces=true}
    ]
\textbf{Observation} —  
\emph{Global Scan:} The series maintains near-zero centering with mean $\mu{=}0.014$ and high variability $\sigma{=}0.712$. The moderate maximum standardized deviation $\max|z|{=}1.89$ indicates well-contained global fluctuations.
\emph{Structural Scan:} Gradient analysis shows moderate trend variation ($\sigma_{\text{grad}}{=}0.0234$), consistent with oscillatory behavior. A strong dominant period emerges at ${\sim}50.0$ time units, suggesting regular cyclical patterns.
\emph{Pattern Scan:} Discord analysis identifies the most atypical subsequence at $t{\approx}2178$ with discord score ${=}0.142$, indicating potential frequency irregularities.

\textbf{Reasoning \& Validation} — I observed that the oscillations within $[2167, 2189]$ exhibit frequency deviation patterns, where the periodic structure changes relative to the expected seasonal baseline. Sliding period analysis ($W{=}120$) reveals typical period $\mu{=}50.2$ with robust $\sigma{=}4.8$. $8$ windows fall outside robust range $[35.8, 64.6]$, confirming frequency drift.

\textbf{Conclusion} — Integrating hierarchical scan insights with targeted frequency deviation validation, the evidence converges on anomalous behavior in interval $[2167, 2189]$. Therefore, the detected anomaly is classified as frequency deviation, precisely localized to $[2167, 2189]$.
    \end{tcolorbox}
    }
    \caption{ExpCoT reasoning trace for frequency deviation anomaly detection.}
    \label{fig:expcot_frequency}
\end{figure}

We adopt the common anomaly taxonomy~\citep{qiu2025tab} with five categories:
(i) \textbf{Out-of-Range / Global Point}, (ii) \textbf{Contextual Point}, (iii) \textbf{Trend Shift},
(iv) \textbf{Seasonal/Frequency Deviation}, and (v) \textbf{Shapelet/Subsequence}.
For each category, we pair characteristic signatures with classical, quantitatively verifiable analyses.
{ExpCoT} is instantiated \emph{per instance} from the ground-truth (GT) anomaly type and temporal annotation, and
follows a disciplined three-stage path:
\textsc{Observation} $\rightarrow$ \textsc{Reasoning \& Validation} $\rightarrow$ \textsc{Conclusion}.
In \emph{Observation}, we perform a unified hierarchical scan of the series: starting with global distributions (e.g., extreme values), then examining structural properties (e.g., trend and periodicity), and finally analyzing localized patterns (e.g., subsequence dissimilarity) to surface candidate anomalies.
\emph{Reasoning \& Validation} aligns the GT type and location with a targeted statistical probe and reports
the resulting numerical evidence.
\emph{Conclusion} integrates these findings into a precise, GT-consistent statement of anomaly type and localization.
Figures~\ref{fig:expcot_global}--\ref{fig:expcot_frequency} illustrate some cases, and we provide the detailed pipeline for each anomaly type below.

\paragraph{(i) Out-of-Range / Global Point.}
\textsc{Observation}:  Apply the defined global–structural–local scan to find salient deviations as candidates for anomaly detection.
\textsc{Reasoning \& Validation}: Apply a $k$-sigma envelope $[\mu-k\sigma,\mu+k\sigma]$ to formalize range
departures; aggregate excursions into contiguous intervals and summarize $(\mu,\sigma)$ and the implied bounds. 
\textsc{Conclusion}:  Retain the GT interval(s) as the definitive localization; envelope breaches serve as
corroborating evidence.

\paragraph{(ii) Contextual Point.}
\textsc{Observation}: Apply the defined global–structural–local scan to find salient deviations as candidates for anomaly detection.
\textsc{Reasoning \& Validation}: Examine fixed-length, $z$-normalized subsequences using the Matrix Profile: let
$d(i)$ be the discord distance and $i^{*}=\arg\max_i d(i)$. Standardize $\{d(i)\}$ to $z(i)$; if
$z(i^{*})>\tau$ (e.g., $\tau{=}3.5$), the subsequence $[i^{*},\,i^{*}{+}m)$ constitutes strong evidence of a
contextual departure. 
\textsc{Conclusion}:  State the GT contextual-point interval(s) as final, summarizing the dominant discord and its
standardized magnitude as quantitative support.

\paragraph{(iii) Trend Shift.}
\textsc{Observation}: Apply the defined global–structural–local scan to find salient deviations as candidates for anomaly detection.
\textsc{Reasoning \& Validation}: Smooth the series and analyze the gradient $g_t$; highlight segments where
$|g_t-\bar g|$ exceeds a multiple of the empirical dispersion of $\{g_t\}$, and merge adjacent exceedances into
candidate intervals indicating a shift in slope or level. 
\textsc{Conclusion}:  Present the GT trend-shift span(s) as the conclusive localization, together with the gradient
summary (center, dispersion, and threshold) as supporting evidence.

\paragraph{(iv) Seasonal/Frequency Deviation.}
\textsc{Observation}:   Apply the defined global–structural–local scan to find salient deviations as candidates for anomaly detection.
\textsc{Reasoning \& Validation}: Estimate the dominant period over sliding windows (FFT-based periodogram) and
identify windows whose periods deviate beyond a robust tolerance around the typical period (e.g., median
$\pm k\times 1.4826\cdot\mathrm{MAD}$). Map these window-level deviations back to the time axis and merge them into
intervals. 
\textsc{Conclusion}:  Declare the GT seasonal/frequency interval(s) as final, reporting the typical period, its robust
dispersion, and the deviation range as quantitative support.

\paragraph{(v) Shapelet/Subsequence.}
\textsc{Observation}:  Apply the defined global–structural–local scan to find salient deviations as candidates for anomaly detection.
\textsc{Reasoning \& Validation}: Use a subsequence dissimilarity scan (e.g., Matrix Profile), prioritizing the most
pronounced discord(s) and, when desired, assessing cross-scale stability across nearby window lengths to strengthen
evidence. 
\textsc{Conclusion}:  When GT specifies a shapelet/subsequence anomaly, return the GT interval(s) as the definitive
localization and include the strongest dissimilar segment(s) as auxiliary evidence.

\paragraph{Instantiation with Ground Truth.}
For every instance, ExpCoT is generated from the GT class and temporal annotation:
\textsc{Observation} anchors on the GT interval(s) and applies the unified scan (global $\rightarrow$ structural
$\rightarrow$ local); \textsc{Reasoning \& Validation} then selects the analysis matched to the GT type and reports
concrete numerical evidence (global envelope deviation, standardized discord magnitude, smoothed-gradient exceedance,
or dominant-period drift); \textsc{Conclusion} integrates these results and retains the GT interval(s) as the
final localization, yielding a faithful, interpretable trace for supervising MLLM training. In practice, these traces are first generated automatically by code to provide quantified validation, and are subsequently refined by human experts for greater fluency and high-fidelity interpretability.

\begin{figure}[h]
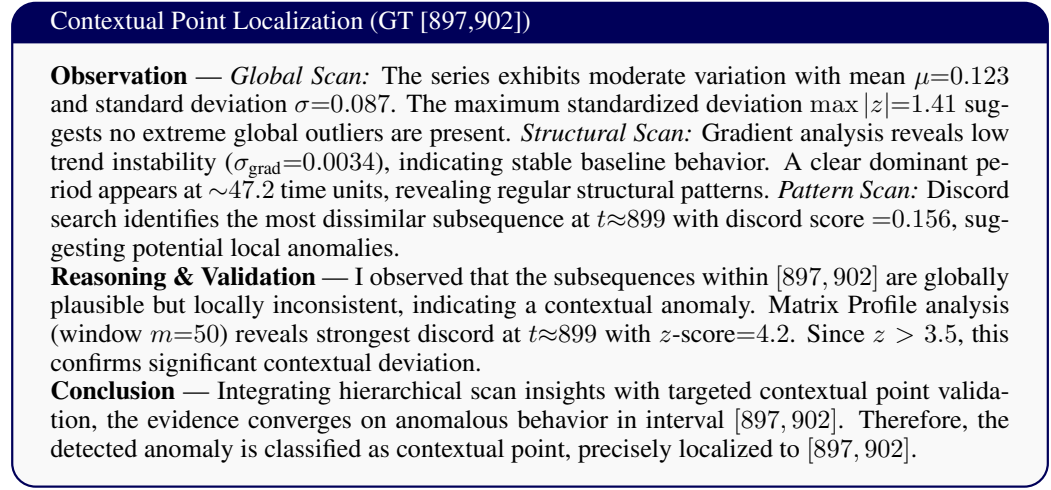

    \centering
    \resizebox{1\linewidth}{!}{
    \begin{tcolorbox}[
      colback=gray!5!white,
      colframe=blue!35!black,
      title={Contextual Point Localization (GT [897,902])},
      boxrule=0.3mm,
      width=\textwidth,
      arc=3mm,
      auto outer arc=true,
      breakable,
      enhanced,
      listing only,
      listing engine=listings,
      listing options={basicstyle=\ttfamily\small, columns=fullflexible, keepspaces=true}
    ]
\textbf{Observation} —  
\emph{Global Scan:} The series exhibits moderate variation with mean $\mu{=}0.123$ and standard deviation $\sigma{=}0.087$. The maximum standardized deviation $\max|z|{=}1.41$ suggests no extreme global outliers are present.
\emph{Structural Scan:} Gradient analysis reveals low trend instability ($\sigma_{\text{grad}}{=}0.0034$), indicating stable baseline behavior. A clear dominant period appears at ${\sim}47.2$ time units, revealing regular structural patterns.
\emph{Pattern Scan:} Discord search identifies the most dissimilar subsequence at $t{\approx}899$ with discord score ${=}0.156$, suggesting potential local anomalies.

\textbf{Reasoning \& Validation} — I observed that the subsequences within $[897, 902]$ are globally plausible but locally inconsistent, indicating a contextual anomaly. Matrix Profile analysis (window $m{=}50$) reveals strongest discord at $t{\approx}899$ with $z$-score${=}4.2$. Since $z > 3.5$, this confirms significant contextual deviation.

\textbf{Conclusion} — Integrating hierarchical scan insights with targeted contextual point validation, the evidence converges on anomalous behavior in interval $[897, 902]$. Therefore, the detected anomaly is classified as a contextual point, precisely localized to $[897, 902]$.
    \end{tcolorbox}
    }
    \caption{Example of ExpCoT reasoning trace for contextual point anomaly detection.}
    \label{fig:expcot_contextual}
\end{figure}

\begin{table}[ht]
\centering
\caption{Comparison of AnomLLM, VisualTimeAnomaly and TSB-UAD.}
\begin{tabular}{lccc}
\toprule
Category & \textbf{AnomLLM} & \textbf{VisualTimeAnomaly} & \textbf{TSB-UAD} \\
\midrule
Synthetic         & \checkmark & \checkmark & \checkmark \\
Real-world        & \xmark     & \checkmark & \checkmark \\
Length range      & 1{,}000    & $\approx$1{,}000 & 1{,}570--230{,}400 \\
Avg. anomaly rate & $\sim$5\%  & $\sim$5--15\% & $\sim$0.2--9.8\% \\
\midrule
\multicolumn{4}{l}{\textbf{Anomaly Types}} \\
\quad Global point      & \checkmark & \checkmark & \checkmark \\
\quad Contextual point  & \checkmark & \checkmark & \checkmark \\
\quad Trend             & \checkmark & \checkmark & \checkmark \\
\quad Seasonal          & \checkmark & \checkmark & \checkmark \\
\quad Shapelet          & \xmark     & \checkmark & \checkmark \\
\bottomrule
\end{tabular}
\label{tab:data}
\end{table}

\section{Experimental  Details} \label{app:exp}
\subsection{Dataset Statistics} \label{app:data}
We evaluate three public resources to assess models’ performance and generalizability across various TSAD scenarios. The detailed dataset statistics and anomaly coverage are summarized in Table~\ref{tab:data}.

1) \textbf{AnomLLM}~\citep{zhou2024can} provides controlled synthetic time-series anomaly detection benchmarks. Following the default generation settings, we generate eight anomaly types: out-of-range, point, frequency, trend, flat-trend, noisy-point, noisy-freq, and noisy-trend. They can be grouped into four categories: range, point, freq, and trend. For nomenclature consistency in this paper, we map the original task names to our taxonomy as follows: Range $\rightarrow$ Global point, Point $\rightarrow$ Contextual point, Freq $\rightarrow$ Seasonal, and Trend $\rightarrow$ Trend. Given this synthetic generation process, {global (out-of-range)} anomalies are typically the easiest to detect, whereas {contextual point}, {trend}, and {seasonal} anomalies are more difficult due to their reliance on local context, regime changes, and frequency shifts, respectively.

2) \textbf{VisualTimeAnomaly}~\citep{xu2025can} converts numerical time series into images across various scenarios; in our study, we focus on the univariate setting and adhere to the default synthetic workflow. The benchmark includes point-wise (global/contextual) and range-wise (trend/seasonal/shapelet) anomalies for univariate series. Within this dataset, {point-wise anomalies} are the hardest to localize visually, whereas {range-wise anomalies} are comparatively easier due to their salient coarse-grained patterns.

3) \textbf{TSB-UAD}~\citep{qiu2025tab} unifies 1{,}635 univariate series from the original TSB-UAD~\citep{paparrizos2022tsb} by filtering out low-quality series (e.g., those without anomalies or with an anomaly ratio $>$10\%), resulting in a high-quality collection that includes both real-world and synthetic datasets. We adopt the official defaults and taxonomy. The TAB-UAD dataset covers both univariate and multivariate settings (treating each multivariate dataset as multiple univariate time series and evaluating them individually). The anomaly coverage includes point (global/contextual) and subsequence (trend/shapelet/seasonal) categories, as well as mixed types. The collected series span diverse domains such as industrial sensors, medical signals, finance, and web traffic, making the benchmark both comprehensive and representative of real-world anomaly detection challenges.

\subsection{Baselines} \label{app:baseline}
For each benchmark, we evaluate three groups of models. For the closed-source MLLMs, we access commercial APIs, including GPT-4o, GPT-4o-mini, Gemini-2.5-Pro, and Gemini-2.5-Flash-Lite. For the open-source counterparts, we rely on HuggingFace checkpoints such as Qwen/Qwen2.5-VL-72b-Instruct and its smaller variants (e.g., 32B/7B/3B). We further compare against supervised fine-tuned baselines, including Qwen2.5-VL-3B-SFT3.2k, fine-tuned on 3,200 instances, and Qwen2.5-VL-3B-SFT32k, fine-tuned on 32,000 instances.

In addition, we include two representative LLM-based temporal reasoning baselines. \textsc{SigLLM}~\citep{alnegheimish2024large}  is a GPT-3.5-based detector for anomaly identification.  We evaluate SigLLM under the default settings provided in its official repository, using the original prompts and raw numerical inputs. \textsc{TimeMaster}~\citep{zhang2025timemaster}, which builds on Qwen2.5-VL-3B with supervised fine-tuning (SFT) and GRPO and adopts image inputs, is also trained under its default public release. For all models except \textsc{SigLLM}, we use the same prompt templates (see Figure~\ref{fig:tsadprompt}) to ensure consistency and fairness.

\subsection{Metrics}
{
We evaluate detection quality using the affiliation-based metrics introduced by \citet{huet2022local}\footnote{The official implementation of these metrics is publicly available, namely \emph{Affi\_Precision}, \emph{Affi\_Recall}, and their harmonic mean \emph{Affi\_F1}. 
These affiliation-based metrics can be viewed as event-level extensions of the classical precision/recall/F1-score to time-series anomaly detection~\citep{huet2022local}. 
Affi\_Precision and Affi\_Recall evaluate each ground-truth event locally, and are parameter-free.
Moreover, their construction via comparison to a random reference predictor makes the resulting scores both theoretically principled and practically useful for TSAD, especially in LLM-based TSAD settings~\citep{zhou2024can, liu2024picture, xu2025can}. 
Below, we provide their detailed definitions.}

{
\textbf{Setup.}
Recall that a univariate time series of length $T$ is denoted by
$\mathbf{X} = \{ \mathbf{x}_t \}_{t=1}^{T}$.
Ground-truth anomaly intervals are given by
\[
  \mathcal{A} 
  \;=\; 
  \bigl\{ (t_s^{(i)}, t_e^{(i)}) \bigr\}_{i=1}^{k},
  \qquad
  1 \le t_s^{(i)} \le t_e^{(i)} \le T ,
\]
where each interval $(t_s^{(i)}, t_e^{(i)})$ denotes the $i$-th anomalous 
segment (with $t_s^{(i)} = t_e^{(i)}$ corresponding to a single-point anomaly).
We assume these intervals are pairwise disjoint.
For convenience, we identify each interval with the corresponding set of time
indices,
\[
  J_i 
  \;=\; 
  \{ t \in \{1,\dots,T\} : t_s^{(i)} \le t \le t_e^{(i)} \}.
\]
Thus the collection of ground-truth events is
\[
  \mathcal{J} = \{ J_j \}_{j=1}^{n},
\]
where $n = k$ and the $J_j$ are pairwise disjoint subsets of 
$\{1, \dots, T\}$.}

{
Similarly, we denote the predicted anomaly intervals by
\[
  \widehat{\mathcal{A}} 
  \;=\; 
  \bigl\{ (\hat{t}_s^{(i)}, \hat{t}_e^{(i)}) \bigr\}_{i=1}^{\hat{k}},
\]
and write
\[
  \widehat{J}_i 
  \;=\; 
  \{ t \in \{1,\dots,T\} : \hat{t}_s^{(i)} \le t \le \hat{t}_e^{(i)} \},
  \qquad
  \widehat{\mathcal{J}} = \{ \widehat{J}_i \}_{i=1}^{\hat{k}}.
\]
All sets $J_j$ and $\widehat{J}_i$ are subsets of the index set
$\mathcal{T} = \{1,\dots,T\}$.
For any $A \subseteq \mathcal{T}$, we write $|A|$ for its cardinality.
For $t \in \mathcal{T}$ and $Y \subseteq \mathcal{T}$, we define
\[
  \operatorname{dist}(t,Y) 
  \;=\; 
  \min_{y \in Y} |t-y|
\]
as the distance (in time indices) from $t$ to the set $Y$, with the convention
that $\operatorname{dist}(t,\varnothing) = +\infty$.}

{
\textbf{Affiliation regions.}
Following~\citep{huet2022local}, we partition the time index set 
$\mathcal{T}$ into \emph{affiliation regions} $\{E_j\}_{j=1}^{n}$, one for
each ground-truth event $J_j$:
\[
  E_j 
  \;=\; 
  \bigl\{ t \in \mathcal{T} : 
          j = \arg\min_{k \in \{1,\dots,n\}} 
          \operatorname{dist}(t, J_k) 
   \bigr\},
\]
with ties broken arbitrarily so that $\{E_j\}_{j=1}^{n}$ forms a partition of
$\mathcal{T}$, i.e.\ $\mathcal{T} = \biguplus_{j=1}^{n} E_j$ and 
$E_j \cap E_k = \varnothing$ for $j \neq k$.}
{
For each $j$, we denote by
\[
  \widetilde{P}_j 
  \;=\; 
  \Bigl(\bigcup_{i=1}^{\hat{k}} \widehat{J}_i\Bigr) \cap E_j
\]
the subset of predicted anomalous time indices that fall inside the affiliation
region $E_j$.}

{
\textbf{Random reference predictor.}
For each $j \in \{1,\dots,n\}$, we define a random reference predictor by
drawing a time index
\[
  X_j \sim \mathrm{Unif}(E_j),
\]
uniformly at random from $E_j$.
The \emph{precision-side baseline distance} is
\[
  D^{\mathrm{prec}}_j 
  \;=\; 
  \operatorname{dist}(X_j, J_j),
\]
and its survival function (complementary CDF) is
\[
  \overline{F}^{\mathrm{prec}}_j(d) 
  \;=\; 
  \mathbb{P}\bigl(D^{\mathrm{prec}}_j \ge d\bigr), 
  \qquad d \ge 0.
\]
Intuitively, $\overline{F}^{\mathrm{prec}}_j(d)$ measures how likely a random
prediction in $E_j$ lies at distance at least $d$ from the true event $J_j$.}

{
For the recall side, for each $j$ and each time index $t \in J_j$, we define
\[
  D^{\mathrm{rec}}_{j,t} 
  \;=\; 
  \operatorname{dist}(t, X_j),
\]
and the corresponding survival function
\[
  \overline{F}^{\mathrm{rec}}_{j,t}(d) 
  \;=\; 
  \mathbb{P}\bigl(D^{\mathrm{rec}}_{j,t} \ge d\bigr),
  \qquad d \ge 0.
\]}

{
\textbf{Affi\_Precision.}
For a fixed ground-truth event $J_j$, the \emph{local affiliation-precision
score} $P_{\mathrm{prec}}(j)$ compares the actual predictions in $E_j$ to the
random baseline:
\[
  P_{\mathrm{prec}}(j)
  \;=\;
  \begin{cases}
    \displaystyle
    \frac{1}{|\widetilde{P}_j|}
    \sum_{t \in \widetilde{P}_j}
      \overline{F}^{\mathrm{prec}}_j
      \bigl(\operatorname{dist}(t, J_j)\bigr),
    & \text{if } |\widetilde{P}_j| > 0, \\[2.2ex]
    \text{(ignored)}, 
    & \text{if } |\widetilde{P}_j| = 0 .
  \end{cases}
\]
Only those events with $|\widetilde{P}_j| > 0$ contribute to the global
precision. Let
\[
  S \;=\; \bigl\{ j \in \{1,\dots,n\} : |\widetilde{P}_j| > 0 \bigr\}
\]
be the set of ground-truth events for which at least some prediction mass
falls into $E_j$. The global \emph{Affi\_Precision} is defined as
\[
  \mathrm{Affi\_Precision}
  \;=\;
  \begin{cases}
    \displaystyle
    \frac{1}{|S|}
    \sum_{j \in S} P_{\mathrm{prec}}(j),
    & \text{if } |S| > 0, \\[1.5ex]
    0, & \text{if } |S| = 0.
  \end{cases}
\]}

{
\textbf{Affi\_Recall.}
For the recall side, each ground-truth event $J_j$ defines a local score
$P_{\mathrm{rec}}(j)$ by averaging, over all time indices $t \in J_j$, how much
better the prediction $\widetilde{P}_j$ is than the random baseline:
\[
  P_{\mathrm{rec}}(j)
  \;=\;
  \frac{1}{|J_j|}
  \sum_{t \in J_j}
    \overline{F}^{\mathrm{rec}}_{j,t}
    \bigl(\operatorname{dist}(t, \widetilde{P}_j)\bigr),
\]
where
\[
  \operatorname{dist}(t, \widetilde{P}_j)
  \;=\; 
  \min_{z \in \widetilde{P}_j} |t - z|,
\]
with the convention that if $\widetilde{P}_j = \varnothing$,
then $\operatorname{dist}(t, \widetilde{P}_j) = +\infty$ and
$\overline{F}^{\mathrm{rec}}_{j,t}(\operatorname{dist}(t, \widetilde{P}_j)) = 0$.}

{
The global \emph{Affi\_Recall} is obtained by averaging
$P_{\mathrm{rec}}(j)$ over all ground-truth events:
\[
  \mathrm{Affi\_Recall}
  \;=\;
  \frac{1}{n}
  \sum_{j = 1}^{n} P_{\mathrm{rec}}(j).
\]}
{
\textbf{Affi\_F1.}
Finally, the \emph{Affi\_F1} score is defined as the harmonic mean of 
Affi\_Precision and Affi\_Recall. Let
\[
  P \;=\; \mathrm{Affi\_Precision}, 
  \qquad
  R \;=\; \mathrm{Affi\_Recall},
\]
then
\[
  \mathrm{Affi\_F1}
  \;=\;
  \begin{cases}
    0, & \text{if } P + R = 0, \\[0.6ex]
    \displaystyle
    \frac{2 P R}{P + R}, & \text{otherwise}.
  \end{cases}
\]}
{
By construction, $\mathrm{Affi\_Precision}$, $\mathrm{Affi\_Recall}$, and
$\mathrm{Affi\_F1}$ all take values in the interval $[0,1]$.}

\subsection{Implementation Details} \label{app:imp}
\paragraph{Time-Series Image Input.}
We follow the common plotting conventions used in prior work on MLLMs~\citep{xu2025can,zhou2024can,zhang2025timemaster} to ensure fairness. The line plots do not include shaded or highlighted regions, and anomalous intervals are not explicitly marked. Each time-series image is rendered at a resolution of $805 \times 124$ pixels.
\begin{table}[ht]
\centering
\caption{Hyperparameter settings.}
\begin{tabular}{lcl|lcl}
\toprule
\textbf{Algorithm} & \textbf{Hyperparameter} & \textbf{Value} & 
\textbf{Algorithm} & \textbf{Hyperparameter} & \textbf{Value} \\
\midrule
\multirow{6}{*}{GRPO} 
  & Max response length & 1024 & \multirow{4}{*}{TimerPO} & $\lambda^{\mathrm{fmt}}$ & 0.1 \\
  & Batch size          & 128  &                          & $\lambda^{\mathrm{cls}}$ & 0.2 \\
  & Mini-batch size     & 128  &                          & $\lambda^{\mathrm{loc}}$ & 0.7 \\
  & KL loss coefficient & 0.001&                          & $\alpha$           & 0.3 \\
  & Group size          & 5    &                          &                    &     \\
  & Learning rate       & 1e-6 &                          &                    &     \\
\bottomrule
\end{tabular}
\label{tab:hyper}
\end{table}

\paragraph{Training Setup.}
We initialize our backbone with the publicly available Qwen2.5-VL-3B-Instruct and Qwen2.5-VL-7B-Instruct checkpoints. Our overall training pipeline only includes a TimerPO stage based purely on reinforcement learning. 
We build our implementation on the public RL training library and the temporal reasoning training framework. 
We summarize our hyperparameter settings in Table~\ref{tab:hyper}, where the GRPO configuration follows \textsc{TimeMaster} for fairness. The models are trained on 3,200 synthetic samples from \textsc{AnomLLM} and evaluated on the \textsc{AnomLLM} synthetic test set, \textsc{VisualTimeAnomaly}, and \textsc{TSB-UAD}, which cover broader anomaly types and varying sequence lengths to assess generalization to unseen real-world scenarios.

\subsection{System Configuration}\label{app:sys}
All experiments were conducted on a computing setup equipped with 4 NVIDIA A100-SXM4 GPUs (80 GB each) and 4 NVIDIA RTX A6000 GPUs (48 GB each) for Qwen-3B, and 4 NVIDIA H100-SXM4 GPUs (96 GB each) for Qwen-7B.

\begin{figure*}[t]
\centering
\includegraphics[width=0.95\textwidth]{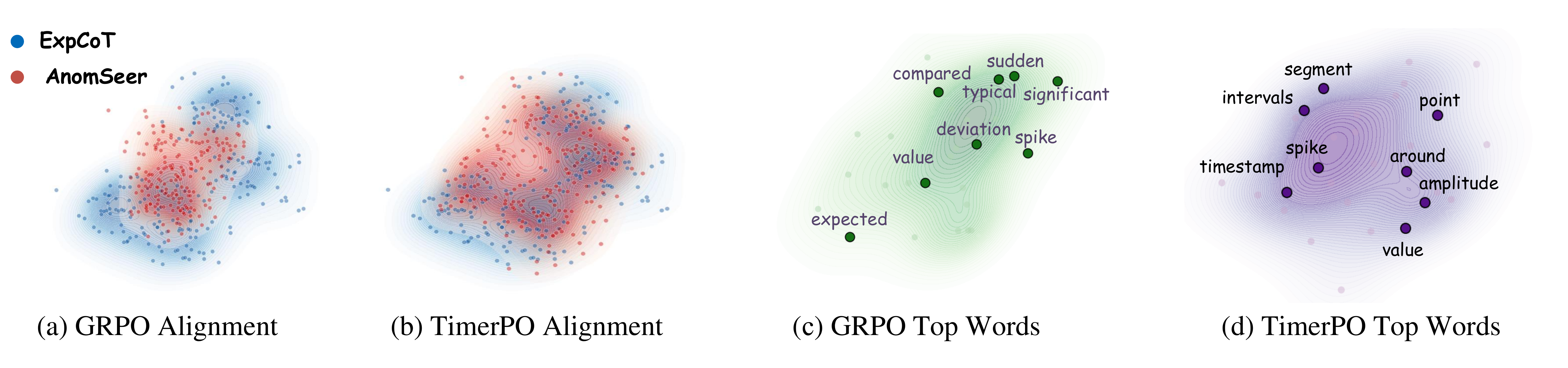}
\vspace{-0.1in}
\caption{Comparison of distributional alignment between ExpCoT (blue) and \methodname{} (red) outputs, along with token usage under GRPO and TimerPO training.}
\label{fig:alignment_grpo}
\end{figure*}

\section{More Experimental Results} \label{app:res}

{\subsection{Confidence Intervals and Computational Cost} \label{app:res_ci}
To complement the results in the main paper, we provide the complete set of performance metrics corresponding to Table~\ref{tab:main_f1}, including mean values over three runs together with their 95\% confidence intervals. As shown in Table~\ref{tab:ci_main}, the consistently small intervals support the robustness of our findings and indicate that \methodname{} performs stably across repeated trials.

We also report the computational profile of \methodname{} (3B) trained on NVIDIA RTX~A6000 GPUs with 48 GB of memory. The training phase requires 12.4 hours of wall-clock time using four GPUs in parallel. For inference, the model operates on a single GPU, utilizing approximately 7 GB of memory and achieving an average latency of 4.8 seconds per time-series sample. These computational characteristics fall within acceptable limits for practical deployment in TSAD scenarios.

\begin{table}[t]
\caption{Mean $\pm$ 95\% confidence interval half-width over 3 seeds.}
\label{tab:ci_main}
\centering
\small
\setlength{\tabcolsep}{5pt}
\renewcommand{\arraystretch}{1.15}
\resizebox{\textwidth}{!}{
\begin{tabular}{
l
c
ccc ccc ccc ccc
c
}
\toprule
\multirow{3}{*}{\textbf{Method}} &
\multicolumn{1}{c}{\textbf{Classification}} &
\multicolumn{13}{c}{\textbf{Location}} \\
\cmidrule(lr){2-2}
\cmidrule(lr){3-15}
& \multirow{2}{*}{\textbf{Accuracy}} &
\multicolumn{3}{c}{\textbf{Frequency}} &
\multicolumn{3}{c}{\textbf{Trend}} &
\multicolumn{3}{c}{\textbf{Range}} &
\multicolumn{3}{c}{\textbf{Point}} &
\multirow{2}{*}{\textbf{Avg F1}} \\
\cmidrule(lr){3-5}
\cmidrule(lr){6-8}
\cmidrule(lr){9-11}
\cmidrule(lr){12-14}
& 
& \textbf{P} & \textbf{R} & \textbf{F1}
& \textbf{P} & \textbf{R} & \textbf{F1}
& \textbf{P} & \textbf{R} & \textbf{F1}
& \textbf{P} & \textbf{R} & \textbf{F1}
& \\
\midrule

\textbf{TimeMaster-3B}
& 57.90$\pm$1.49
& 57.30$\pm$1.24 & 50.30$\pm$0.25 & 51.40$\pm$0.50
& 76.00$\pm$1.24 & 77.30$\pm$0.25 & 76.60$\pm$1.24
& 77.80$\pm$1.24 & 83.50$\pm$0.25 & 80.10$\pm$0.25
& 77.70$\pm$1.24 & 82.10$\pm$0.25 & 79.60$\pm$1.24
& 71.92$\pm$0.81
\\

\textbf{ANOMSEER-3B}
& 62.80$\pm$1.24
& 63.70$\pm$1.24 & 58.40$\pm$1.24 & 58.90$\pm$1.24
& 84.20$\pm$0.50 & 85.90$\pm$0.25 & 84.90$\pm$0.25
& 83.30$\pm$0.75 & 89.20$\pm$0.25 & 85.60$\pm$0.25
& 86.00$\pm$0.25 & 90.30$\pm$0.25 & 87.80$\pm$0.25
& 79.30$\pm$0.50
\\

\textbf{ANOMSEER-7B}
& 65.00$\pm$1.24
& 68.30$\pm$1.24 & 59.40$\pm$0.50 & 60.80$\pm$0.50
& 86.60$\pm$1.24 & 89.00$\pm$1.24 & 87.70$\pm$1.24
& 91.60$\pm$0.25 & 97.80$\pm$0.99 & 94.30$\pm$0.25
& 93.40$\pm$0.25 & 96.90$\pm$1.24 & 94.90$\pm$0.25
& 84.42$\pm$0.56
\\

\bottomrule
\end{tabular}}

\end{table}

\begin{figure}[t]
\centering
\includegraphics[width=0.95\textwidth]{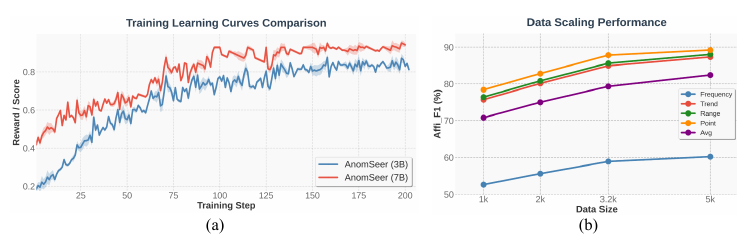}
\caption{(a) Learning curves of training score versus training steps for the 3B and 7B models, and (b) data-scaling performance for the 3B model evaluated from 1k to 5k training examples.}
\label{fig:learning}
\end{figure}

{
\subsection{Learning Curves \&  Data Scaling}
We present the learning curves and data-scaling results in Figure~\ref{fig:learning}. We first observe that the learning curves for both the 3B and 7B models exhibit stable and monotonic improvement, with performance rising rapidly during the initial 50-100 training steps before gradually stabilizing. In addition, scaling the training set from 1k to 5k examples yields consistent gains across all four tasks. The average Affinity F1 score continues to improve as the dataset grows, with no clear signs of saturation. These results suggest that the current data regime remains in a growth phase, and further increasing the amount of training data is likely to yield additional performance gains.
}

{\subsection{Optimization and Alignment Ablation}
\label{app:ablation}
{
\paragraph{Advantage-level orthogonalization vs.\ gradient-level projection.}
We compared TimerPO to two multi-objective optimization baselines: (i) a weighted-sum objective (no projection) and (ii) PCGrad-style gradient orthogonalization~\citep{yu2020gradient}. Table~\ref{tab:integrated_ablation} summarizes the results. TimerPO consistently outperforms both weighted-sum and gradient-level projection across all anomaly types. Orthogonalizing auxiliary signals at the advantage level promotes complementary contributions prior to gradient computation, whereas PCGrad only modifies gradients when explicit conflicts are detected. This reduces partial interference between objectives and results in smoother, lower-variance optimization trajectories.

\vspace{0.5em}
\begin{table}[h!]
\caption{Comparison of orthogonalization strategies (top) and alignment objectives (bottom).}
\label{tab:integrated_ablation}
\centering
\small
\begin{tabular}{lccccc}
\hline
\textbf{Method} & \textbf{Freq.} & \textbf{Trend} & \textbf{Range} & \textbf{Point} & \textbf{Avg} \\
\hline
\multicolumn{6}{l}{\textit{Orthogonalization strategies}} \\
TimerPO (ours)              & 58.9 & 84.9 & 85.6 & 87.8 & \textbf{79.3} \\
Weighted-sum (no proj.)     & 53.5 & 81.1 & 83.5 & 85.4 & 75.9 \\
PCGrad (gradient level)     & 54.2 & 80.2 & 84.5 & 86.4 & 76.3 \\
\hline
\multicolumn{6}{l}{\textit{Alignment objectives}} \\
TimerPO (ours)              & 58.9 & 84.9 & 85.6 & 87.8 & \textbf{79.3} \\
Cosine similarity           & 42.2 & 73.8 & 84.1 & 86.8 & 71.7 \\
CLIP-style similarity       & 48.5 & 74.1 & 84.3 & 86.7 & 73.4 \\
\hline
\end{tabular}

\end{table}
\vspace{0.5em}

\paragraph{Replacing OT with cosine or contrastive similarity.}
To isolate the contribution of OT-based alignment, the OT module in TimerPO was replaced with two alternatives: (i) token-wise cosine similarity and (ii) a CLIP-style InfoNCE objective (temperature $=0.07$).
As shown in Table~\ref{tab:integrated_ablation}, OT yields a 6.8$\%$ improvement in average F1 relative to cosine and contrastive similarity. OT provides structure-aware alignment by modeling semantic distances between reasoning tokens rather than treating tokens independently. These findings indicate that OT geometry plays an essential role in aligning model reasoning with temporally structured anomaly patterns.
}

\subsection{Extensibility}
\label{app:extensibility}

We now discuss the extensibility of the proposed method, with results summarized in Tab.~\ref{tab:extensibility}.
\textbf{1) Multivariate time series.} Although the main experiments focus on univariate data, the framework is not limited to this setting. Multivariate inputs can be converted into a unified visual representation by rendering each variable as a subplot within a single image. Empirical results on a multivariate benchmark demonstrate that the method generalizes effectively beyond the univariate setting.
\textbf{2) Short-term and boundary anomalies.}
Short-duration or boundary anomalies are typically underrepresented in existing datasets and therefore challenging to detect reliably. A simple targeted augmentation strategy yields notable improvements on a dedicated evaluation set of such cases. These findings indicate that lightweight preprocessing and sampling adjustments can enhance robustness in challenging anomaly scenarios.
}

\begin{table}[h!]
\caption{Results for multivariate evaluation (top) and short/boundary anomaly robustness (bottom).}
\label{tab:extensibility}
\centering
\small
\begin{tabular}{lcccc}
\toprule
\textbf{Setting} 
& \textbf{GPT-4o} 
& \textbf{Gemini-2.5} 
& \textbf{Qwen2.5-VL} 
& \textbf{Ours (7B)} \\
\midrule
Multivariate (synthetic / real-world)
& 62.7 / 54.2 
& 77.3 / 65.2 
& 45.0 / 24.5 
& \textbf{83.5 / 72.4} \\
\midrule
Short / boundary anomalies
& \multicolumn{2}{c}{Original: 56.2} 
& \multicolumn{2}{c}{\textbf{+ Augmentation: 75.8}} \\
\bottomrule
\end{tabular}

\end{table}

}
{
\subsection{Comparison with Traditional TSAD Methods}}
\label{app:traditional}

{
To provide a unified view of classical time-series anomaly detection (TSAD) methods and our framework, Table~\ref{tab:traditional_full} summarizes representative baselines across four anomaly types. Traditional approaches such as FFT, Matrix Profile, gradient-based detection, ARIMA, and statistical thresholding operate directly on raw signals and typically produce detection outputs only. While they can perform well in specific scenarios, they often rely on careful parameter tuning (e.g., window selection or differencing) and exhibit limited robustness across diverse anomaly patterns.}

{
In contrast, \methodname{} outperforms traditional approaches across all anomaly types and, more importantly, supports detection, classification, and natural language reasoning within a single model. This broader output capability enables interpretability and generalization across diverse anomaly patterns, rather than optimizing for a single metric or domain-specific signal property.
}

\begin{table}[h!]
\caption{Comparison with classical TSAD baselines and the proposed \methodname{}.}
\label{tab:traditional_full}
\centering
\small
\begin{tabular}{lcccccc}
\toprule
\textbf{Method} &
\textbf{Capability} &
\textbf{Freq.} &
\textbf{Trend} &
\textbf{Range} &
\textbf{Point} &
\textbf{Avg} \\
\midrule
FFT                                  & Location only & 65.9 & 18.0 & 28.5 & 27.8 & 35.1 \\
Matrix Profile                        & Location only & 11.4 & 29.4 & 67.2 & 87.4 & 48.9 \\
Gradient                              & Location only & 57.1 & 58.5 & 55.8 & 65.9 & 59.3 \\
Ensemble (voting)                     & Location only & 59.0 & 17.4 & 69.2 & 92.4 & 59.5 \\
ARIMA                                 & Location only & 62.6 & 4.9  & 67.3 & 74.2 & 52.3 \\
Thresholding                          & Location only & 57.6 & 25.2 & 60.0 & 47.6 & 47.6 \\
\midrule
\methodname{} (3B)                    & Loc. + Cls. + Reasoning & 58.9 & 84.9 & 85.6 & 87.8 & \textbf{79.3} \\
\methodname{} (7B)                    & Loc. + Cls. + Reasoning & 60.8 & 87.7 & 94.3 & 94.9 & \textbf{84.4} \\
\bottomrule
\end{tabular}

\end{table}

\subsection{Details on Effect of TimerPO}\label{app:res1}
To highlight the advantage of TimerPO over vanilla GRPO in temporal reasoning, we further compare their behaviors in Figure~\ref{fig:alignment_grpo}. While GRPO narrows the distributional gap to some extent, the model outputs remain relatively constrained and still exhibit a clear mismatch compared to expert reasoning. A similar pattern is evident in token usage: GRPO-trained outputs are dominated by outcome-oriented words such as compared and expected, whereas TimerPO encourages the use of more fine-grained, temporally grounded terms like timestamp, intervals, and amplitude, which anchor reasoning to concrete temporal structures. These findings confirm that TimerPO provides a more principled enhancement over GRPO, enabling models to move beyond surface outcome alignment toward genuine temporal reasoning.

\subsection{More Case Studies on Reasoning}

We provide several case studies illustrating our model's complete reasoning process on corresponding data visualizations. These examples show our approach's effectiveness in focusing on specific segments and timestamps for fine-grained analysis.
We also present a failure case in Fig.~\ref{fig:fail}: a short anomaly within the interval [998, 1000] at the sequence's very end goes undetected by \methodname{}, which incorrectly classifies it as 'normal'. This highlights the need for future work to improve the sensitivity of MLLMs to such boundary-case anomalies.

\begin{figure}[h!]
\begin{AIbox}[{Response of \methodname{} on the ground-truth trend anomaly [[800, 850]].}]
\label{fig:case1}
{
    \textbf{Input Image}: 
    \begin{center}
        \includegraphics[width=1\textwidth]{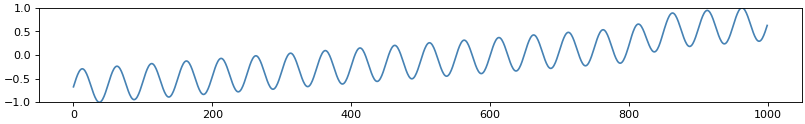}
    \end{center}

    \textbf{\methodname{} Response}: 
\textcolor{deepgreen}{<think>}
The plot shows a general wave-like pattern with intervals where the amplitude becomes slightly elevated.
Such behavior could hint at a potential trend change, but the deviations are not very pronounced.
The global structure remains smooth, with no sudden spikes or sharp dips that would qualify as clear anomalies.
However, I did notice a small irregularity forming around timestep 850, where the local pattern begins to diverge slightly from the surrounding trend. The fluctuation is modest in amplitude, with peaks rising only marginally higher and troughs dipping slightly deeper than the preceding cycles. This subtle change suggests a weak signal of trend variation rather than a strong anomaly.
\textcolor{deepgreen}{</think>}

\textcolor{deepblue}{<answer>}
[[800, 900]]
\textcolor{deepblue}{</answer>}

\textcolor{deeppurple}{<class>}
trend
\textcolor{deeppurple}{</class>}

}
\end{AIbox}

\end{figure}

\begin{figure}[h!]
\begin{AIbox}[{Response of \methodname{} on the contextual point anomaly [[200,231],[622,673]]}]
{
    \textbf{Input Image}: 
    \begin{center}
        \includegraphics[width=1\textwidth]{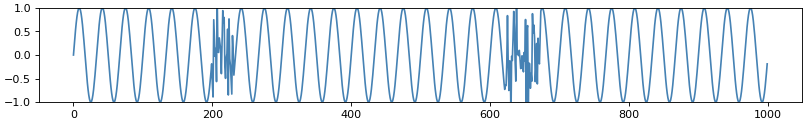}
    \end{center}

    \textbf{\methodname{} Response}: 
\textcolor{deepgreen}{<think>}
The time series exhibits two noticeable intervals that deviate from the surrounding pattern. The first occurs between roughly 200 and 250, while the second spans around 600 to 700. Both segments rise sharply compared with the smoother baseline, resembling local spikes rather than gradual changes. These departures stand out in their immediate neighborhoods but do not exceed the overall global range of the series. Because these deviations break the contextual continuity of the waveform, they are better characterized as contextual point anomalies.
\textcolor{deepgreen}{</think>}

\textcolor{deepblue}{<answer>}
[[200, 250], [600, 700]]
\textcolor{deepblue}{</answer>}

\textcolor{deeppurple}{<class>}
contextual point
\textcolor{deeppurple}{</class>}

}
\end{AIbox}
\label{fig:case2}
\end{figure}

\begin{figure}[h!]
\begin{AIbox}[{Response of \methodname{} on the global point anomaly [[358,380],[910,931]]}]
{
    \textbf{Input Image}: 
    \begin{center}
        \includegraphics[width=1\textwidth]{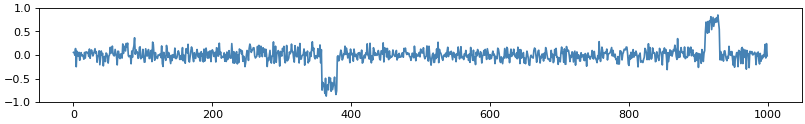}
    \end{center}

    \textbf{\methodname{} Response}: 
\textcolor{deepgreen}{<think>}
The plot contains several sharp, discontinuous excursions in amplitude.
By scanning the sequence for abrupt changes that break this continuity, I identified three distinct intervals: one near 350, another around 880, and a third close to 900. Each spike pops outside the local band, then quickly snaps back.
These intervals deviate noticeably from the surrounding baseline, suggesting global point anomalies in the series.
\textcolor{deepgreen}{</think>}

\textcolor{deepblue}{<answer>}
[[350, 450], [880, 950]]
\textcolor{deepblue}{</answer>}

\textcolor{deeppurple}{<class>}
global point
\textcolor{deeppurple}{</class>}

}
\end{AIbox}
\label{fig:case3}
\end{figure}

\begin{figure}[h!]
\begin{AIbox}[{Response of \methodname{} on the freq anomaly [[653,676],[811,846]]}]
{
    \textbf{Input Image}: 
    \begin{center}
        \includegraphics[width=1\textwidth]{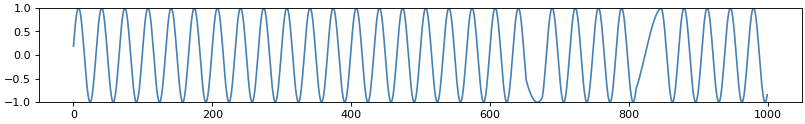}
    \end{center}

    \textbf{\methodname{} Response}: 
\textcolor{deepgreen}{<think>}
To identify anomalies, I first examine the trend and periodic structure of the plot.  
The early section displays a regular sinusoidal rhythm with no unusual deviations.  
Around the 650th timestep, however, the frequency begins to distort, showing dips and irregular oscillations that break the expected cycle.  
In addition, I observed another disturbance near the 850th mark, where the pattern again deviates from the baseline.  
This suggests a change in the underlying dynamics, with the presence of a seasonal anomaly disrupting the expected pattern. 
\textcolor{deepgreen}{</think>}

\textcolor{deepblue}{<answer>}
[[580, 700], [850, 900]]
\textcolor{deepblue}{</answer>}

\textcolor{deeppurple}{<class>}
seasonal
\textcolor{deeppurple}{</class>}

}
\end{AIbox}
\label{fig:case4}
\end{figure}

\begin{figure}[t!]
\begin{AIboxDark}[{Failure case where \methodname{} fails to detect the ground-truth trend anomaly at [998,1000]}]
{
    \textbf{Input Image}: 
    \begin{center}
        \includegraphics[width=1\textwidth]{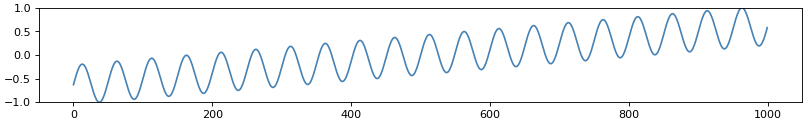}
    \end{center}

    \textbf{\methodname{} Response}: 
\textcolor{deepgreen}{<think>}
No anomalies detected in the plot.
\textcolor{deepgreen}{</think>}

\textcolor{deepblue}{<answer>}
[]
\textcolor{deepblue}{</answer>}

\textcolor{deeppurple}{<class>}
normal
\textcolor{deeppurple}{</class>}

}
\end{AIboxDark}
\caption{Failure case}
\label{fig:fail}
\end{figure}

\section{Extended Related Work}
\textbf{LLM-based time series anomaly detection (TSAD)} is an emerging area, with several exploratory methods recently proposed. These approaches vary in modalities, backbones, and integration strategies. For example, SigLLM~\citep{alnegheimish2024large} and CoLLaTe~\citep{chen2025synergizing} use numeric–text inputs with GPT-3.5 or GPT-4, relying on prompting with external post-processing or task-specific TSAD modules, but without reasoning ability. LLMAD~\citep{liu2025large} augments GPT-4-turbo with retrieval-based domain knowledge to support anomaly classification and localization, though it requires an external database for prompting. On the multimodal side, TAMA~\citep{zhuang2024see} and VLM4TS~\citep{he2025harnessing} employ image–text inputs with GPT-4o, together with post-processing or ViT-based components. More recently, Time-RA~\citep{yang2025time} applies large-scale SFT on Qwen2.5-VL-7B, but its coverage remains incomplete, particularly in anomaly localization.  In contrast, our method uses a compact open-source backbone, Qwen2.5-VL-3B/7B, and reinforcement learning to directly equip the model with anomaly classification, localization, and reasoning, without external modules or proprietary APIs.

{
\textbf{Optimal Transport} (OT) offers a principled geometric framework for aligning probability distributions and has seen increasing adoption in both reinforcement learning (RL) and large language model (LLM) alignment. In RL, OT has been leveraged to structure learning signals and align task or policy distributions~\citep{klink2022curriculum, asadulaev2024rethinking, chen2020sequence}, notably in curriculum design~\citep{klink2022curriculum} and as a regularizer for offline policy learning~\citep{asadulaev2024rethinking}. Within LLM, OT has been used to support preference modeling~\citep{li2025optimal, melnyk2024distributional, desideri2012multiple, xu2026stackelberg, li2025aplot}, by aligning full reward distributions~\citep{melnyk2024distributional} or applying token-level weighting schemes to highlight semantically important regions~\citep{li2025optimal}. Most of these approaches focus on final outcome alignment, operating over entire sequences or aggregated behaviors. But our work applies OT at the reasoning-token level, aligning the model’s intermediate reasoning steps with structured ExpCoT traces derived from classical TSAD primitives. This enables process-level supervision, enhancing the model’s temporal reasoning capabilities rather than merely refining output preferences.
}

{
\textbf{Multi-objective optimization} methods~\citep{desideri2012multiple, yu2020gradient, liu2021conflict, wei2024mmpareto} aim to stabilize training across competing tasks by projecting conflicting gradients into compatible directions. For example, PCGrad~\citep{yu2020gradient} explicitly projects one task’s gradient onto the normal plane of another when conflicts arise. In contrast, our TimerPO introduces orthogonal projection in the {advantage space}, not to resolve inter-task interference, but to preserve the independent contribution of an {auxiliary reasoning advantage}. Since this auxiliary signal reflects structured supervision rather than a separate objective, our projection design allows it to complement the main anomaly detection reward without disruption. To the best of our knowledge, this is the first approach to combine token-level OT alignment with advantage-space disentanglement to enhance temporal reasoning in multimodal LLMs.
}
\end{document}

%% file: math_commands.tex
\usepackage{amsmath,amsfonts,bm}

\def\eqref#1{equation~\ref{#1}}

\def\1{\bm{1}}

\DeclareMathAlphabet{\mathsfit}{\encodingdefault}{\sfdefault}{m}{sl}
\SetMathAlphabet{\mathsfit}{bold}{\encodingdefault}{\sfdefault}{bx}{n}













%% file: icml2026_paper.bbl
\begin{thebibliography}{57}
\providecommand{\natexlab}[1]{#1}
\providecommand{\url}[1]{\texttt{#1}}
\expandafter\ifx\csname urlstyle\endcsname\relax
  \providecommand{\doi}[1]{doi: #1}\else
  \providecommand{\doi}{doi: \begingroup \urlstyle{rm}\Url}\fi

\bibitem[Alnegheimish et~al.(2024)Alnegheimish, Nguyen, Berti-Equille, and Veeramachaneni]{alnegheimish2024large}
Alnegheimish, S., Nguyen, L., Berti-Equille, L., and Veeramachaneni, K.
\newblock Large language models can be zero-shot anomaly detectors for time series?
\newblock \emph{arXiv preprint arXiv:2405.14755}, 2024.

\bibitem[Asadulaev et~al.(2024)Asadulaev, Korst, Korotin, Egiazarian, Filchenkov, and Burnaev]{asadulaev2024rethinking}
Asadulaev, A., Korst, R., Korotin, A., Egiazarian, V., Filchenkov, A., and Burnaev, E.
\newblock Rethinking optimal transport in offline reinforcement learning.
\newblock \emph{Advances in Neural Information Processing Systems}, 37:\penalty0 123592--123607, 2024.

\bibitem[Bai et~al.(2025)Bai, Chen, Liu, Wang, Ge, Song, Dang, Wang, Wang, Tang, et~al.]{bai2025qwen2}
Bai, S., Chen, K., Liu, X., Wang, J., Ge, W., Song, S., Dang, K., Wang, P., Wang, S., Tang, J., et~al.
\newblock Qwen2. 5-vl technical report.
\newblock \emph{arXiv preprint arXiv:2502.13923}, 2025.

\bibitem[Bhatnagar et~al.(2021)Bhatnagar, Kassianik, Liu, Lan, Yang, Cassius, Sahoo, Arpit, Subramanian, Woo, Saha, Jagota, Gopalakrishnan, Singh, Krithika, Maddineni, Cho, Zong, Zhou, Xiong, Savarese, Hoi, and Wang]{bhatnagar2021merlion}
Bhatnagar, A., Kassianik, P., Liu, C., Lan, T., Yang, W., Cassius, R., Sahoo, D., Arpit, D., Subramanian, S., Woo, G., Saha, A., Jagota, A.~K., Gopalakrishnan, G., Singh, M., Krithika, K.~C., Maddineni, S., Cho, D., Zong, B., Zhou, Y., Xiong, C., Savarese, S., Hoi, S., and Wang, H.
\newblock Merlion: A machine learning library for time series.
\newblock 2021.

\bibitem[Bonneel et~al.(2011)Bonneel, Van De~Panne, Paris, and Heidrich]{bonneel2011displacement}
Bonneel, N., Van De~Panne, M., Paris, S., and Heidrich, W.
\newblock Displacement interpolation using lagrangian mass transport.
\newblock In \emph{Proceedings of the 2011 SIGGRAPH Asia conference}, pp.\  1--12, 2011.

\bibitem[Caffarelli \& McCann(2010)Caffarelli and McCann]{caffarelli2010free}
Caffarelli, L.~A. and McCann, R.~J.
\newblock Free boundaries in optimal transport and monge-ampere obstacle problems.
\newblock \emph{Annals of mathematics}, pp.\  673--730, 2010.

\bibitem[Chen et~al.(2025)Chen, Zhang, Pang, Zimmermann, and Deng]{chen2025synergizing}
Chen, F., Zhang, L., Pang, G., Zimmermann, R., and Deng, S.
\newblock Synergizing large language models and task-specific models for time series anomaly detection.
\newblock \emph{arXiv preprint arXiv:2501.05675}, 2025.

\bibitem[Chen et~al.(2020)Chen, Bai, Tao, Zhang, Wang, Wang, Henao, and Carin]{chen2020sequence}
Chen, L., Bai, K., Tao, C., Zhang, Y., Wang, G., Wang, W., Henao, R., and Carin, L.
\newblock Sequence generation with optimal-transport-enhanced reinforcement learning.
\newblock In \emph{Proceedings of the AAAI Conference on Artificial Intelligence}, volume~34, pp.\  7512--7520, 2020.

\bibitem[Cuturi(2013)]{cuturi2013sinkhorn}
Cuturi, M.
\newblock Sinkhorn distances: Lightspeed computation of optimal transport.
\newblock \emph{Advances in neural information processing systems}, 26, 2013.

\bibitem[D{\'e}sid{\'e}ri(2012)]{desideri2012multiple}
D{\'e}sid{\'e}ri, J.-A.
\newblock Multiple-gradient descent algorithm (mgda) for multiobjective optimization.
\newblock \emph{Comptes Rendus Mathematique}, 350\penalty0 (5-6):\penalty0 313--318, 2012.

\bibitem[Feng et~al.(2025)Feng, Xue, Liu, and An]{feng2025group}
Feng, L., Xue, Z., Liu, T., and An, B.
\newblock Group-in-group policy optimization for llm agent training.
\newblock \emph{arXiv preprint arXiv:2505.10978}, 2025.

\bibitem[Gao et~al.(2024)Gao, Koker, Queen, Hartvigsen, Tsiligkaridis, and Zitnik]{gao2024units}
Gao, S., Koker, T., Queen, O., Hartvigsen, T., Tsiligkaridis, T., and Zitnik, M.
\newblock Units: A unified multi-task time series model.
\newblock \emph{Advances in Neural Information Processing Systems}, 37:\penalty0 140589--140631, 2024.

\bibitem[Goldstein \& Dengel(2012)Goldstein and Dengel]{goldstein2012histogram}
Goldstein, M. and Dengel, A.
\newblock Histogram-based outlier score (hbos): A fast unsupervised anomaly detection algorithm.
\newblock \emph{KI-2012: poster and demo track}, 1:\penalty0 59--63, 2012.

\bibitem[Goswami et~al.(2024)Goswami, Szafer, Choudhry, Cai, Li, and Dubrawski]{goswami2024moment}
Goswami, M., Szafer, K., Choudhry, A., Cai, Y., Li, S., and Dubrawski, A.
\newblock Moment: A family of open time-series foundation models.
\newblock \emph{arXiv preprint arXiv:2402.03885}, 2024.

\bibitem[Guo et~al.(2025)Guo, Yang, Zhang, Song, Zhang, Xu, Zhu, Ma, Wang, Bi, et~al.]{guo2025deepseek}
Guo, D., Yang, D., Zhang, H., Song, J., Zhang, R., Xu, R., Zhu, Q., Ma, S., Wang, P., Bi, X., et~al.
\newblock Deepseek-r1: Incentivizing reasoning capability in llms via reinforcement learning.
\newblock \emph{arXiv preprint arXiv:2501.12948}, 2025.

\bibitem[He et~al.(2025)He, Alnegheimish, and Reimherr]{he2025harnessing}
He, Z., Alnegheimish, S., and Reimherr, M.
\newblock Harnessing vision-language models for time series anomaly detection.
\newblock \emph{arXiv preprint arXiv:2506.06836}, 2025.

\bibitem[Huet et~al.(2022)Huet, Navarro, and Rossi]{huet2022local}
Huet, A., Navarro, J.~M., and Rossi, D.
\newblock Local evaluation of time series anomaly detection algorithms.
\newblock In \emph{Proceedings of the 28th ACM SIGKDD Conference on Knowledge Discovery and Data Mining}, pp.\  635--645, 2022.

\bibitem[Hyndman \& Athanasopoulos(2018)Hyndman and Athanasopoulos]{hyndman2018forecasting}
Hyndman, R.~J. and Athanasopoulos, G.
\newblock \emph{Forecasting: principles and practice}.
\newblock OTexts, 2018.

\bibitem[Klink et~al.(2022)Klink, Yang, D’Eramo, Peters, and Pajarinen]{klink2022curriculum}
Klink, P., Yang, H., D’Eramo, C., Peters, J., and Pajarinen, J.
\newblock Curriculum reinforcement learning via constrained optimal transport.
\newblock In \emph{International Conference on Machine Learning}, pp.\  11341--11358. PMLR, 2022.

\bibitem[Kong et~al.(2026)Kong, Yang, Wang, Liu, Liang, Jin, Zohren, Pei, Liu, and Wen]{kong2026achievingtimeseriesreasoning}
Kong, Y., Yang, Y., Wang, S., Liu, C., Liang, Y., Jin, M., Zohren, S., Pei, D., Liu, Y., and Wen, Q.
\newblock Achieving time series reasoning requires rethinking model design, tasks formulation, and evaluation, 2026.
\newblock URL \url{https://arxiv.org/abs/2502.01477}.

\bibitem[Li et~al.(2025{\natexlab{a}})Li, Huzhang, Zhang, Wang, and Zeng]{li2025optimal}
Li, M., Huzhang, G., Zhang, H., Wang, X., and Zeng, A.
\newblock Optimal transport-based token weighting scheme for enhanced preference optimization.
\newblock \emph{arXiv preprint arXiv:2505.18720}, 2025{\natexlab{a}}.

\bibitem[Li et~al.(2024)Li, Chen, Chai, and Xiong]{li2024gilot}
Li, X., Chen, J., Chai, Y., and Xiong, H.
\newblock Gilot: Interpreting generative language models via optimal transport.
\newblock In \emph{Forty-first International Conference on Machine Learning}, 2024.

\bibitem[Li et~al.(2025{\natexlab{b}})Li, Feng, Guo, Hu, Gao, and Wan]{li2025aplot}
Li, Z., Feng, Y., Guo, D., Hu, J., Gao, A., and Wan, X.
\newblock Aplot: Robust reward modeling via adaptive preference learning with optimal transport.
\newblock In \emph{Proceedings of the 2025 Conference on Empirical Methods in Natural Language Processing}, pp.\  5524--5538, 2025{\natexlab{b}}.

\bibitem[Liu et~al.(2021)Liu, Liu, Jin, Stone, and Liu]{liu2021conflict}
Liu, B., Liu, X., Jin, X., Stone, P., and Liu, Q.
\newblock Conflict-averse gradient descent for multi-task learning.
\newblock \emph{Advances in Neural Information Processing Systems}, 34:\penalty0 18878--18890, 2021.

\bibitem[Liu et~al.(2008)Liu, Ting, and Zhou]{liu2008isolation}
Liu, F.~T., Ting, K.~M., and Zhou, Z.-H.
\newblock Isolation forest.
\newblock In \emph{2008 eighth ieee international conference on data mining}, pp.\  413--422. IEEE, 2008.

\bibitem[Liu et~al.(2024)Liu, Liu, and Prakash]{liu2024picture}
Liu, H., Liu, C., and Prakash, B.~A.
\newblock A picture is worth a thousand numbers: Enabling llms reason about time series via visualization.
\newblock \emph{arXiv preprint arXiv:2411.06018}, 2024.

\bibitem[Liu et~al.(2025{\natexlab{a}})Liu, Zhang, Qian, Ma, Qin, Bansal, Lin, Rajmohan, and Zhang]{liu2025large}
Liu, J., Zhang, C., Qian, J., Ma, M., Qin, S., Bansal, C., Lin, Q., Rajmohan, S., and Zhang, D.
\newblock Large language models can deliver accurate and interpretable time series anomaly detection.
\newblock In \emph{Proceedings of the 31st ACM SIGKDD Conference on Knowledge Discovery and Data Mining V. 2}, pp.\  4623--4634, 2025{\natexlab{a}}.

\bibitem[Liu et~al.(2025{\natexlab{b}})Liu, Han, Yu, Li, and You]{liu2025time}
Liu, Z., Han, P., Yu, H., Li, H., and You, J.
\newblock Time-r1: Towards comprehensive temporal reasoning in llms.
\newblock \emph{arXiv preprint arXiv:2505.13508}, 2025{\natexlab{b}}.

\bibitem[Luo et~al.(2025)Luo, Zhou, Cheng, Wang, Wang, Pan, and Zhang]{luo2025time}
Luo, Y., Zhou, Y., Cheng, M., Wang, J., Wang, D., Pan, T., and Zhang, J.
\newblock Time series forecasting as reasoning: A slow-thinking approach with reinforced llms.
\newblock \emph{arXiv preprint arXiv:2506.10630}, 2025.

\bibitem[Melnyk et~al.(2024)Melnyk, Mroueh, Belgodere, Rigotti, Nitsure, Yurochkin, Greenewald, Navratil, and Ross]{melnyk2024distributional}
Melnyk, I., Mroueh, Y., Belgodere, B., Rigotti, M., Nitsure, A., Yurochkin, M., Greenewald, K., Navratil, J., and Ross, J.
\newblock Distributional preference alignment of llms via optimal transport.
\newblock \emph{Advances in Neural Information Processing Systems}, 37:\penalty0 104412--104442, 2024.

\bibitem[Paparrizos et~al.(2022)Paparrizos, Kang, Boniol, Tsay, Palpanas, and Franklin]{paparrizos2022tsb}
Paparrizos, J., Kang, Y., Boniol, P., Tsay, R.~S., Palpanas, T., and Franklin, M.~J.
\newblock Tsb-uad: an end-to-end benchmark suite for univariate time-series anomaly detection.
\newblock \emph{Proceedings of the VLDB Endowment}, 15\penalty0 (8):\penalty0 1697--1711, 2022.

\bibitem[Park et~al.(2018)Park, Hoshi, and Kemp]{park2018multimodal}
Park, D., Hoshi, Y., and Kemp, C.~C.
\newblock A multimodal anomaly detector for robot-assisted feeding using an lstm-based variational autoencoder.
\newblock \emph{IEEE Robotics and Automation Letters}, 3\penalty0 (3):\penalty0 1544--1551, 2018.

\bibitem[Qiu et~al.(2025)Qiu, Li, Qiu, Hu, Zhou, Wu, Li, Guo, Zhou, Sheng, et~al.]{qiu2025tab}
Qiu, X., Li, Z., Qiu, W., Hu, S., Zhou, L., Wu, X., Li, Z., Guo, C., Zhou, A., Sheng, Z., et~al.
\newblock Tab: Unified benchmarking of time series anomaly detection methods.
\newblock \emph{arXiv preprint arXiv:2506.18046}, 2025.

\bibitem[Ren et~al.(2019)Ren, Xu, Wang, Yi, Huang, Kou, Xing, Yang, Tong, and Zhang]{ren2019time}
Ren, H., Xu, B., Wang, Y., Yi, C., Huang, C., Kou, X., Xing, T., Yang, M., Tong, J., and Zhang, Q.
\newblock Time-series anomaly detection service at microsoft.
\newblock In \emph{Proceedings of the 25th ACM SIGKDD international conference on knowledge discovery \& data mining}, pp.\  3009--3017, 2019.

\bibitem[Sch{\"o}lkopf et~al.(1999)Sch{\"o}lkopf, Williamson, Smola, Shawe-Taylor, and Platt]{scholkopf1999support}
Sch{\"o}lkopf, B., Williamson, R.~C., Smola, A., Shawe-Taylor, J., and Platt, J.
\newblock Support vector method for novelty detection.
\newblock \emph{Advances in neural information processing systems}, 12, 1999.

\bibitem[Shao et~al.(2024)Shao, Wang, Zhu, Xu, Song, Bi, Zhang, Zhang, Li, Wu, et~al.]{shao2024deepseekmath}
Shao, Z., Wang, P., Zhu, Q., Xu, R., Song, J., Bi, X., Zhang, H., Zhang, M., Li, Y., Wu, Y., et~al.
\newblock Deepseekmath: Pushing the limits of mathematical reasoning in open language models.
\newblock \emph{arXiv preprint arXiv:2402.03300}, 2024.

\bibitem[Shentu et~al.(2024)Shentu, Li, Zhao, Shu, Rao, Pan, Yang, and Guo]{shentu2024towards}
Shentu, Q., Li, B., Zhao, K., Shu, Y., Rao, Z., Pan, L., Yang, B., and Guo, C.
\newblock Towards a general time series anomaly detector with adaptive bottlenecks and dual adversarial decoders.
\newblock \emph{arXiv preprint arXiv:2405.15273}, 2024.

\bibitem[Sutton \& Barto(2018)Sutton and Barto]{sutton2018reinforcement}
Sutton, R.~S. and Barto, A.~G.
\newblock \emph{Reinforcement learning: An introduction}.
\newblock MIT press, 2018.

\bibitem[Tan et~al.(2025)Tan, Merrill, Gottesman, Althoff, Evans, and Hartvigsen]{tan2025inferring}
Tan, M., Merrill, M.~A., Gottesman, Z., Althoff, T., Evans, D., and Hartvigsen, T.
\newblock Inferring events from time series using language models.
\newblock \emph{arXiv preprint arXiv:2503.14190}, 2025.

\bibitem[Thill et~al.(2017)Thill, Konen, and B{\"a}ck]{thill2017time}
Thill, M., Konen, W., and B{\"a}ck, T.
\newblock Time series anomaly detection with discrete wavelet transforms and maximum likelihood estimation.
\newblock In \emph{Intern. Conference on Time Series (ITISE)}, volume~2, pp.\  11--23, 2017.

\bibitem[Villani et~al.(2008)]{villani2008optimal}
Villani, C. et~al.
\newblock \emph{Optimal transport: old and new}, volume 338.
\newblock Springer, 2008.

\bibitem[Wei \& Hu(2024)Wei and Hu]{wei2024mmpareto}
Wei, Y. and Hu, D.
\newblock Mmpareto: Boosting multimodal learning with innocent unimodal assistance.
\newblock \emph{arXiv preprint arXiv:2405.17730}, 2024.

\bibitem[Wei et~al.(2025)Wei, Duchenne, Copet, Carbonneaux, Zhang, Fried, Synnaeve, Singh, and Wang]{wei2025swe}
Wei, Y., Duchenne, O., Copet, J., Carbonneaux, Q., Zhang, L., Fried, D., Synnaeve, G., Singh, R., and Wang, S.~I.
\newblock Swe-rl: Advancing llm reasoning via reinforcement learning on open software evolution.
\newblock \emph{arXiv preprint arXiv:2502.18449}, 2025.

\bibitem[Wu et~al.(2025)Wu, Qiu, Li, Wang, Hu, Guo, Xiong, and Yang]{wu2024catch}
Wu, X., Qiu, X., Li, Z., Wang, Y., Hu, J., Guo, C., Xiong, H., and Yang, B.
\newblock {CATCH}: Channel-aware multivariate time series anomaly detection via frequency patching.
\newblock In \emph{ICLR}, 2025.

\bibitem[Xie et~al.(2024)Xie, Li, He, Xu, Wen, Zhang, Chen, Shi, and Pei]{xie2024chatts}
Xie, Z., Li, Z., He, X., Xu, L., Wen, X., Zhang, T., Chen, J., Shi, R., and Pei, D.
\newblock Chatts: Aligning time series with llms via synthetic data for enhanced understanding and reasoning.
\newblock \emph{arXiv preprint arXiv:2412.03104}, 2024.

\bibitem[Xu et~al.(2026)Xu, Zhang, Jia, and Jin]{xu2026stackelberg}
Xu, C., Zhang, Z., Jia, T., and Jin, Y.
\newblock Stackelberg self-annotation: A robust approach to data-efficient llm alignment.
\newblock \emph{Advances in Neural Information Processing Systems}, 38:\penalty0 62912--62949, 2026.

\bibitem[Xu et~al.(2021)Xu, Wu, Wang, and Long]{xu2021anomaly}
Xu, J., Wu, H., Wang, J., and Long, M.
\newblock Anomaly transformer: Time series anomaly detection with association discrepancy.
\newblock \emph{arXiv preprint arXiv:2110.02642}, 2021.

\bibitem[Xu et~al.(2025)Xu, Wang, Liang, Yu, Zhao, and Shu]{xu2025can}
Xu, X., Wang, H., Liang, Y., Yu, P.~S., Zhao, Y., and Shu, K.
\newblock Can multimodal llms perform time series anomaly detection?
\newblock \emph{arXiv preprint arXiv:2502.17812}, 2025.

\bibitem[Yang et~al.(2025)Yang, Liu, Song, Ying, Wang, Bamford, Vyetrenko, Bian, and Wen]{yang2025time}
Yang, Y., Liu, Z., Song, L., Ying, K., Wang, Z., Bamford, T., Vyetrenko, S., Bian, J., and Wen, Q.
\newblock Time-ra: Towards time series reasoning for anomaly with llm feedback.
\newblock \emph{arXiv preprint arXiv:2507.15066}, 2025.

\bibitem[Yeh et~al.(2016)Yeh, Zhu, Ulanova, Begum, Ding, Dau, Silva, Mueen, and Keogh]{yeh2016matrix}
Yeh, C.-C.~M., Zhu, Y., Ulanova, L., Begum, N., Ding, Y., Dau, H.~A., Silva, D.~F., Mueen, A., and Keogh, E.
\newblock Matrix profile i: all pairs similarity joins for time series: a unifying view that includes motifs, discords and shapelets.
\newblock In \emph{2016 IEEE 16th international conference on data mining (ICDM)}, pp.\  1317--1322. Ieee, 2016.

\bibitem[Yu et~al.(2020)Yu, Kumar, Gupta, Levine, Hausman, and Finn]{yu2020gradient}
Yu, T., Kumar, S., Gupta, A., Levine, S., Hausman, K., and Finn, C.
\newblock Gradient surgery for multi-task learning.
\newblock \emph{Advances in neural information processing systems}, 33:\penalty0 5824--5836, 2020.

\bibitem[Zhang et~al.(2025{\natexlab{a}})Zhang, Liu, Qiu, Liu, Pei, Wang, and Long]{zhang2025timesbert}
Zhang, H., Liu, Y., Qiu, Y., Liu, H., Pei, Z., Wang, J., and Long, M.
\newblock Timesbert: A bert-style foundation model for time series understanding.
\newblock \emph{arXiv preprint arXiv:2502.21245}, 2025{\natexlab{a}}.

\bibitem[Zhang et~al.(2025{\natexlab{b}})Zhang, Feng, Guo, Wu, Dong, and Xu]{zhang2025timemaster}
Zhang, J., Feng, L., Guo, X., Wu, Y., Dong, Y., and Xu, D.
\newblock Timemaster: Training time-series multimodal llms to reason via reinforcement learning.
\newblock \emph{arXiv preprint arXiv:2506.13705}, 2025{\natexlab{b}}.

\bibitem[Zhou et~al.(2023)Zhou, Niu, Sun, Jin, et~al.]{zhou2023one}
Zhou, T., Niu, P., Sun, L., Jin, R., et~al.
\newblock One fits all: Power general time series analysis by pretrained lm.
\newblock \emph{Advances in neural information processing systems}, 36:\penalty0 43322--43355, 2023.

\bibitem[Zhou \& Yu(2024)Zhou and Yu]{zhou2024can}
Zhou, Z. and Yu, R.
\newblock Can llms understand time series anomalies?
\newblock \emph{arXiv preprint arXiv:2410.05440}, 2024.

\bibitem[Zhuang et~al.(2024)Zhuang, Yan, Zhang, Wang, Zhang, and Gu]{zhuang2024see}
Zhuang, J., Yan, L., Zhang, Z., Wang, R., Zhang, J., and Gu, Y.
\newblock See it, think it, sorted: Large multimodal models are few-shot time series anomaly analyzers.
\newblock \emph{arXiv preprint arXiv:2411.02465}, 2024.

\bibitem[Zong et~al.(2018)Zong, Song, Min, Cheng, Lumezanu, Cho, and Chen]{zong2018deep}
Zong, B., Song, Q., Min, M.~R., Cheng, W., Lumezanu, C., Cho, D., and Chen, H.
\newblock Deep autoencoding gaussian mixture model for unsupervised anomaly detection.
\newblock In \emph{ICLR}, 2018.

\end{thebibliography}
